\documentclass[final,5p,times,twocolumn]{elsarticle}
\biboptions{authoryear}
\usepackage{amsmath}
\usepackage{algorithmicx,algorithm}
\usepackage{soul}
\usepackage{bm}
\usepackage{natbib}
\usepackage{amssymb}
\usepackage{array}
\usepackage{tabularx}
\usepackage{pifont}
\usepackage{multirow}
\usepackage{booktabs}
\usepackage{wrapfig}
\usepackage{tabularx}
\usepackage{newfloat}
\usepackage{listings}
\usepackage{subcaption}
\usepackage{graphicx}
\usepackage{subcaption}

\soulregister\cite7
\soulregister\citep7
\soulregister\ref7

\usepackage{xcolor}
\definecolor{ccr}{RGB}{52,146,207}
\usepackage{hyperref}
\hypersetup{
	colorlinks=true,
	citecolor=ccr, 
	linkcolor=ccr, 
	urlcolor=ccr
}
\usepackage{booktabs}
\usepackage{tabularx}
\usepackage{caption}
\usepackage[nameinlink]{cleveref}

\DeclareCaptionLabelFormat{bold}{\textbf{#1 #2}}
\crefname{figure}{Fig.}{Figs.}
\crefname{equation}{Eq.}{Eqs.}
\crefname{table}{Table}{Tables}

\usepackage{amssymb}

\makeatletter
\AtBeginDocument{\def\@citecolor{ccr}}
\AtBeginDocument{\def\@linkcolor{ccr}}
\AtBeginDocument{\def\@urlcolor{ccr}}
\makeatother

\journal{Expert Systems With Applications}
\begin{document}

\begin{frontmatter}



\title{VGTS: Visually Guided Text Spotting for Novel Categories in Historical Manuscripts}


\author[1,2,3]{Wenbo Hu}
\ead{wenbo@stu.ecnu.edu.cn}
\author[1,2]{Hongjian Zhan}
\ead{ecnuhjzhan@foxmail.com}
\author[1,2]{Xinchen Ma}
\ead{xcma@stu.ecnu.edu.cn}
\author[4]{Cong Liu}
\ead{congliu2@iflytek.com}
\author[4]{Bing Yin}
\ead{bingyin@iflytek.com}
\author[1,2]{Yue Lu\corref{mycorrespondingauthor}}
\ead{ylu@cs.ecnu.edu.cn}
\author[3]{Ching Y. Suen}
\ead{chingyee.suen@concordia.ca}
\cortext[mycorrespondingauthor]{Corresponding author}

\address[1]{School of Communication and Electronic Engineering, East China Normal University, Shanghai, 200062, China}
\address[2]{Shanghai Key Laboratory of Multidimensional Information	Processing, Shanghai, 200241, China}
\address[3]{Centre for Pattern Recognition and Machine Intelligence, Concordia University, Montreal, H3G 1M8, Canada}
\address[4]{iFLYTEK Research, iFLYTEK, Hefei, 230088, China}

\begin{abstract}
In the field of historical manuscript research, scholars frequently encounter novel symbols in ancient texts, investing considerable effort in their identification and documentation. Although existing object detection methods achieve impressive performance on known categories, they struggle to recognize novel symbols without retraining. To address this limitation, we propose a Visually Guided Text Spotting (VGTS) approach that accurately spots novel characters using just one annotated support sample. The core of VGTS is a spatial alignment module consisting of a Dual Spatial Attention (DSA) block and a Geometric Matching (GM) block. The DSA block aims to identify, focus on, and learn discriminative spatial regions in the support and query images, mimicking the human visual spotting process. It first refines the support image by analyzing inter-channel relationships to identify critical areas, and then refines the query image by focusing on informative key points. The GM block, on the other hand, establishes the spatial correspondence between the two images, enabling accurate localization of the target character in the query image. To tackle the example imbalance problem in low-resource spotting tasks, we develop a novel torus loss function that enhances the discriminative power of the embedding space for distance metric learning. To further validate our approach, we introduce a new dataset featuring ancient Dongba hieroglyphics (DBH) associated with the Naxi minority of China. Extensive experiments on the DBH dataset and other public datasets, including EGY, VML-HD, TKH, and NC, show that VGTS consistently surpasses state-of-the-art methods. The proposed framework exhibits great potential for application in historical manuscript text spotting, enabling scholars to efficiently identify and document novel symbols with minimal annotation effort.
\end{abstract}




\end{frontmatter}


\section{Introduction}
To preserve cultural heritage, archives create digital libraries by scanning or photographing historical manuscripts \citep{DIG01}. Human summarization remains essential in deciphering and categorizing these ancient texts, which often consist of handwritten characters. Researchers employ Optical Character Recognition (OCR) techniques \citep{OCR01} to facilitate manuscript digitization. However, the digitization of historical manuscripts faces significant challenges due to inherent limitations:

\textbf{(1) Open-set problem.} Most methods perform well on trained categories but falter when encountering novel characters, which is a common scenario in historical manuscripts \citep{OPEN}. This challenge is exacerbated as researchers often discover new categories while examining ancient texts, as detailed in the appendix. While unsupervised methods address this challenge, their effectiveness is limited \citep{DSW,UN1,UN2}.

\textbf{(2) Long-tailed distribution.} The reliance on extensive annotated data is problematic in this domain, given that historical manuscripts typically feature a limited number of pages and scarce training samples. The long-tailed distribution of characters in ancient manuscripts worsens data sparsity, with some characters appearing only once \citep{LT01}. For instance, in the TKH dataset \citep{TKH}, out of 1492 categories, 436 appeared only once.

Moreover, traditional methods necessitate text lines as input, requiring time-consuming and error-prone preprocessing, layout analysis, and character segmentation \citep{LINE01, LINE02}. While recent efforts focus on page-level text spotting \citep{2step}, they still face challenges like error propagation and the need for substantial training data \citep{CD01, CD02}. Additionally, certain historical characters, such as Dongba hieroglyphics or notary signs \citep{NOTARY}, are not compatible with most OCR methods \citep{SEQ05, SEQ06,SEQ07} due to the absence in standard input method editors, further complicating the textualization of results.

Recently, cognitive research has demonstrated that humans possess a remarkable ability to recognize objects when provided with guidance \citep{BRAIN}, as exemplified by human visual searches in scenes with the aid of literacy cards, among other contextual cues. In this process, humans heavily rely on knowledge of the supporting images and contextual information. This dependence on support images (\textit{e.g. literacy cards}) is believed to stem from the prefrontal cortex and is projected to lower-level visual cortex structures. Support information acts as a reference, directing attention to specific visual features. Meanwhile, spatial context information narrows the search area, guiding attention towards more relevant locations. The integration of both sources of information results in the identification of the region of interest.  

To preserve culture and better manage historical manuscripts, we propose a one-shot learning model inspired by human learning cognitive research. One-shot learning focuses on learning patterns of a specific category and generalizing them to unseen categories using only a single annotated example, instead of numerous annotated samples. Our model takes an antique book page image and an example image of the desired character class, spotting all characters in the manuscript, even if the support image belongs to an unseen category. To address example imbalance, we introduce a novel torus loss function, enhancing the distance metric's embedding space discriminability. Our approach handles varying characters and symbols, such as hieroglyphics, Arabic, Chinese, and notary signs in historical manuscripts, offering a flexible and efficient solution for text spotting. Our contributions can be summarized as follows:

(1) We propose a flexible text spotting model that accurately and reliably identifies novel characters with just one annotated example, eliminating the need for additional fine-tuning or retraining.

(2) We propose an innovative dual spatial attention block, which extracts discriminative features from the support set to direct the model's focus to crucial regions in the query image. We also introduce a novel 'torus loss', designed to concentrate on challenging examples and to develop a more effective distance metric for text spotting in low-resource settings.

(3) We contribute a new dataset of ancient Dongba script manuscripts, comprised of hieroglyphic characters. 

\section{Related work}
\subsection{Text Spotting}
Certain segmentation-based methods \citep{SB01, SB02, SB03} assume that the dataset is already partitioned into text lines or words, thereby rendering these methods more akin to an image retrieval task, where the goal is to retrieve the most relevant images in the dataset. Additionally, some text spotting techniques typically detect each text instance using a trained detector, then identify the cropped text region using a sequence decoder \citep{TWO01, TWO02}. However, such a strategy undermines performance robustness since the detection and recognition phases are separated, and the error from the recognition model cannot be used to optimize the detector. In recent times, there has been growing interest in end-to-end scene text detection and recognition, which offer the benefit of achieving both detection and recognition tasks by sharing the acquired features between the two steps \citep{CD01, CD02, CD03, CD04, CD05}. Nevertheless, these methods usually require an extensive corpus of data to enable adequate training and are unable to identify new categories. \citet{DSW} using a sliding window scheme with deep features to accomplish the text spotting task is also one of the solutions. This approach can be seen as a kind of unsupervised learning, so there is no open-set problem, unfortunately, this sliding window based approach usually does not achieve the best results. \citet{QBS} propose a word spotting method based on the ``query-by-string" approach, which allows the user to retrieve all instances of a given string in a document image. However, while this method can be effective for seen characters, it does not address the challenge of mining unseen characters, which is particularly relevant for ancient texts. 

Furthermore, many ancient texts do not yet have a complete lexicon, and consequently, there are no corresponding input method editors, making it challenging to identify and retrieve specific strings in these documents. Historical manuscripts often pose a significant challenge due to their limited availability of annotated images. For instance, the Pascal VOC dataset \citep{VOC}, widely utilized in few-shot object detection, includes a meager 1.6 categories and 2.9 instances per image on average. In stark contrast, the data distribution of historical manuscript images is significantly broader, as evidenced by the TKH dataset, which boasts an average of 72.5 categories and 323.8 character instances per image. Drawing inspiration from few-shot object detection and to address the challenge of limited data in historical manuscripts, \citet{FSCD01} design a few-shot character recognition method that aims to overcome the low-resource problem and enable the model to recognize unseen alphabets. Nonetheless, \citet{FSCD01, FSCD02} are only applicable to text line images, necessitating prior layout analysis to segment the original page into lines. In contrast, our proposed method can process page images without requiring additional layout analysis.
\begin{figure*}[!t]
	\begin{center}
		\includegraphics[clip,width=15cm]{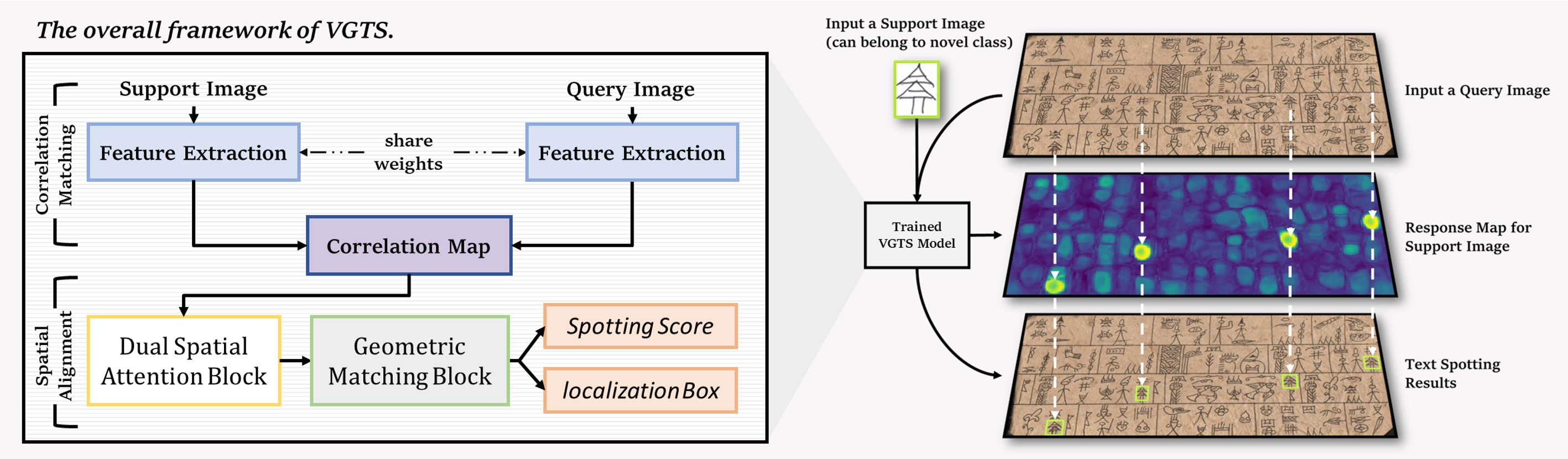}
	\end{center}
	\vspace{-0.5cm}
	\renewcommand{\figurename}{\textbf{Fig.}}
	\caption{The overall framework of VGTS. The proposed framework can spot novel interested character categories conditioned with one support image. Based on the feature extracted by the backbone network, the correlation matching module computes the correlation map to match the pair of individual feature maps, while the spatial alignment module predicts the localization boxes and spotting results.}
	\label{fig:PL}
	\vspace{-0.5cm}
\end{figure*}

\subsection{Few-shot Object Detection}
In the realm of few-shot learning, extent techniques have predominantly been designed for classification tasks. Optimization-based methods effectively assimilate novel categories via gradient-based optimization on a paltry number of annotated samples \citep{OSCO01, OSCO03, OSCO04}. Conversely, the metric-based learning paradigm seeks to glean a common feature space and demarcate categories based on a distance metric \citep{OSCM01, OSCM03}. One-shot learning \citep{ONES02,ONES01} stands as an extreme manifestation of the few-shot learning paradigm, characterized by each novel encountered category boasting a singular labeled example.  

The goal of few-shot object detection is to discern the location and identity of objects in query images with the aid of a limited set of labeled support images \citep{OS04, OS02, OS03, OS09, OS06, OS05, OS10, OS11}. To circumvent the open-set problem, a two-stage approach is often employed, in which the model is first trained on base categories and then fine-tuned on a small number of samples for novel categories. However, this approach can lead to confusion between novel and base categories, resulting in decreased model efficacy as the number of novel categories increases \citep{OS03, OS09}. In situations where training samples are scarce, some approaches may have to struggle to locate regions of interest for novel objects, particularly those with unlearnable shape priors and fine-tuned RPNs \citep{RPN}. However, certain methods have demonstrated the ability to detect new categories without fine-tuning, such as OS2D \citep{OS02}, which uses a similarity map and learned dense matches to establish correspondence between support and query images. This model excels at learning pixel-matching relations between the support and query images, thereby allowing it to handle novel classes without fine-tuning. Our proposed model is also based on dense correlation matching features. However, in few-shot tasks with a small number of samples, attention must be explicitly directed toward the target item (based on the support image) since most visual prediction models exhibit free-viewing behavior. Compared with OS2D, our model design strategy is closer to the human cognitive process, despite the different objectives of our task. Additionally, we discuss the spotting process in low-resource scenarios and propose a new loss function for it. CoAE \citep{OS04} is also a method that does not require fine-tuning, proposing a non-local RPN to tackle the problem of one-shot object detection. However, RPN-based methods often use a predefined set of anchor boxes with different scales and aspect ratios to generate region proposals, which may not be able to capture the minute variations in size and aspect ratio of small objects, especially in historical manuscripts.

\section{Proposed method}
By providing only a single sample of each character type in the support image gallery, the model can effectively locate the characters within the manuscript. As illustrated in \cref{fig:PL}, the proposed model comprises two modules: the correlation matching module and the spatial alignment module. In the correlation matching module, a feature extractor initially extracts the features of both the query and support images. The relationship between feature map pairs is then computed to obtain the correlation map. Subsequently, the spatial alignment module learns the transformation parameters that enable the model to localize the support image within the query image. Since certain characters in ancient manuscripts cannot yet be inputted using an input editor, we select items from the support image gallery $S_{gallery} \in C_{base} \cup C_{novel}$ and paste them onto a blank image of the same size as the query image $I_{q} \in T_{test}$. This creates a new reference image to be displayed as the result of text spotting, where $C_{base}$ denotes the base category that appears during the training process, and $C_{novel}$ denotes the novel category that does not appear during training.

\subsection{Correlation Matching}
\subsubsection{Feature Extractor}
For the input support and query images, $(I_{s}, I_{q})$, are passed through two feature extraction CNN branches with shared weights.
Since most of the antique manuscripts are insufficient to train the network from scratch, the feature extractor can only gain experience from the base category. It cannot get targeted training from the novel category. For each input support image $I_s^{i}$, the feature map $f_{\hat{s}}^{i} \in R^{h_{\hat{s}}^{i} \times w_{\hat{s}}^{i} \times D}$ obtained after backbone, is uniformly resized to the same size $f_{\hat{s}} \rightarrow f_{s} \in R^{h_{s} \times w_{s} \times D}$. 
Finally, the feature extraction networks extract the support feature $F(I_{s})=f_{s} \in R^{h_{s} \times w_{s} \times D}$ and query feature $F(I_{q})=f_{q} \in R^{h_{q} \times w_{q} \times D}$, respectively, where $F$ represents the backbone, $h \times w$ is the spatial resolution and $D$ is the channels.

\subsubsection{Correlation Map}
To ensure the model can have sufficient generalizability to unseen categories. Intuitively, the model must fully avail of the reference information provided by the support image. A common strategy for computing the correlation map is to match a pair of individual feature maps using normalized cosine similarity \citep{COS01, COS02}. Given a pair of individual features $f_{s} \in R^{h_{s} \times w_{s} \times D}$ and $f_{q} \in R^{h_{q} \times w_{q} \times D}$ of support and query images, the correlation map is computed as:
\begin{equation}
	C_{abkl} = \phi(f_{q},f_{s}) = \frac{f_{q} \cdot f_{s}^{T}}{\Vert f_{q} \Vert \Vert f_{s} \Vert} \in R^{h_{q} \times w_{q} \times h_{s} \times w_{s}},
\end{equation}
where $\phi(\cdot)$ means cosine similarity, and $C_{abkl}$ denotes the matching score between the $(a,b)$-th position in query feature map and $(k,l)$-th position in support feature map. The result is thus a 4D tensor that captures the similarities between all pairs of spatial locations. 
\begin{figure}[!t]
	\begin{center}
		\includegraphics[clip,width=8.5cm]{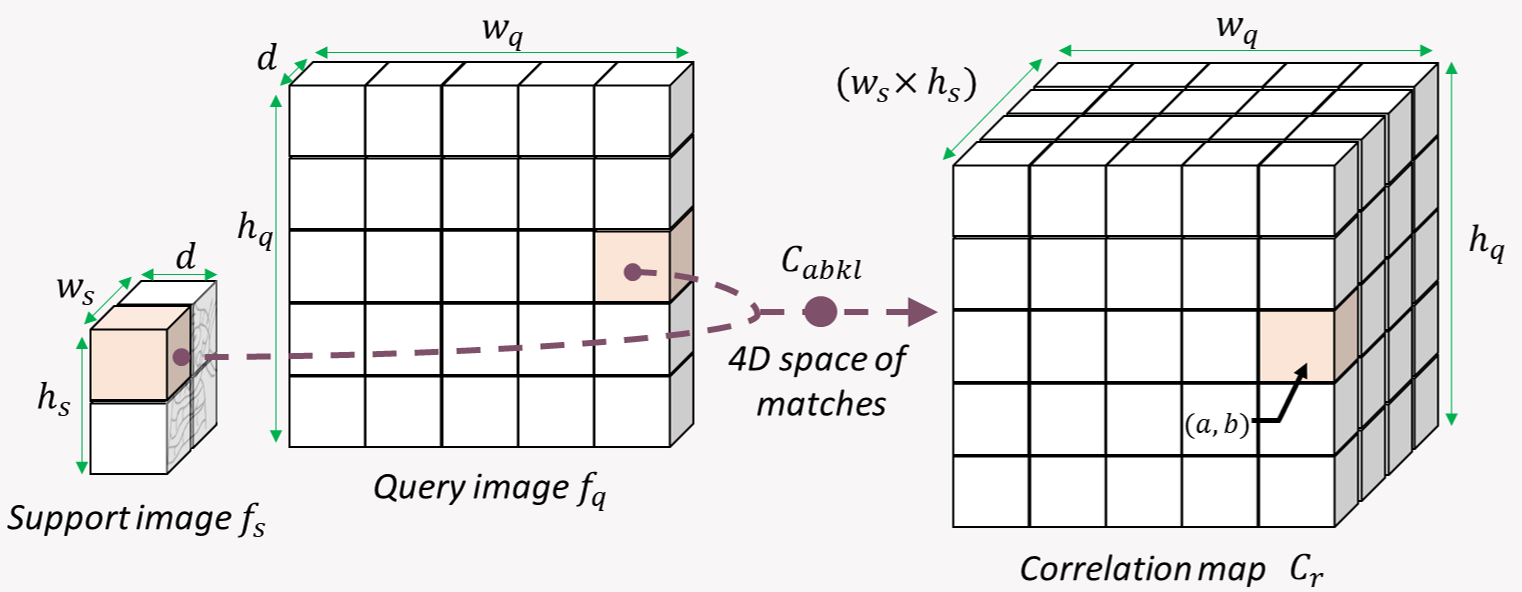} 
	\end{center}
	\vspace{-0.5cm}
	\renewcommand{\figurename}{\textbf{Fig.}}
	\caption{Correlation map computation using pairs of individual features involves first extracting image descriptors $f_{s}$ and $f_{q}$ from images $I_{s}$ and $I_{q}$, respectively. Subsequently, all pairs of individual feature matches, $f_{s}^{kl}$ and $f_{q}^{ab}$, are represented in the 4D space of matches $(a,b,k,l)$, where the matching score is stored in a 4D correlation tensor. This 4D tensor is then reshaped into a 3D tensor with dimensions $h_{q}$, $w_{q}$, and $(h_{s}\times w_{s})$, allowing for the construction of a correlation map. At a specific spatial location $(a,b)$, the correlation map $C_{r}$ provides an aggregation of all similarities between $f_{q}(a,b)$ and all $f_{s}$.}
	\label{fig:CM}
	\vspace{-0.5cm}
\end{figure}
\subsection{Spatial Alignment}
The spatial alignment module consists of two blocks, the dual spatial attention (DSA) block and the geometric matching (GM) block. 
The primary purpose of the dual spatial attention block is to find, focus and learn discriminative spatial regions on support/query images. And the primary purpose of the geometric matching block is to find the spatial mapping relationship between the support image and the query image. These two blocks stimulate each other, which helps to get more accurate text spotting results.

Unlike the geometric matching methods \citep{GM01, GM02, GM03} that have been utilized for correspondence learning, the number of characters that appear in the antique manuscripts is very large, the character shapes are usually tiny, and the character-to-character constructions are very similar. For this reason, we develop a spatial alignment module to facilitate locating the position of the character instance given by the support image on the query image at the dense pixel level. Specifically, the spatial alignment module aims to find spatial correspondence between a pair of support and query images. To achieve this, the obtained 4-dimensional correlation map, can be reshaped to a 3-dimensional tensor with dimensions $h_{q}$,$w_{q}$ and $(h_{s} \times w_{s})$, \textit{i.e.}, $C_{r} = R^{h_{q} \times w_{q} \times (h_{s} \times w_{s}) }$. 
The reshaped correlation map $C_{r}$ can be considered as a dense $h_{q} \times w_{q}$ grid with $(h_{s} \times w_{s})$-dimensional local features as shown in \cref{fig:CM}. 
\begin{figure*}[!h]
	\begin{center}
		\includegraphics[clip,width=15cm]{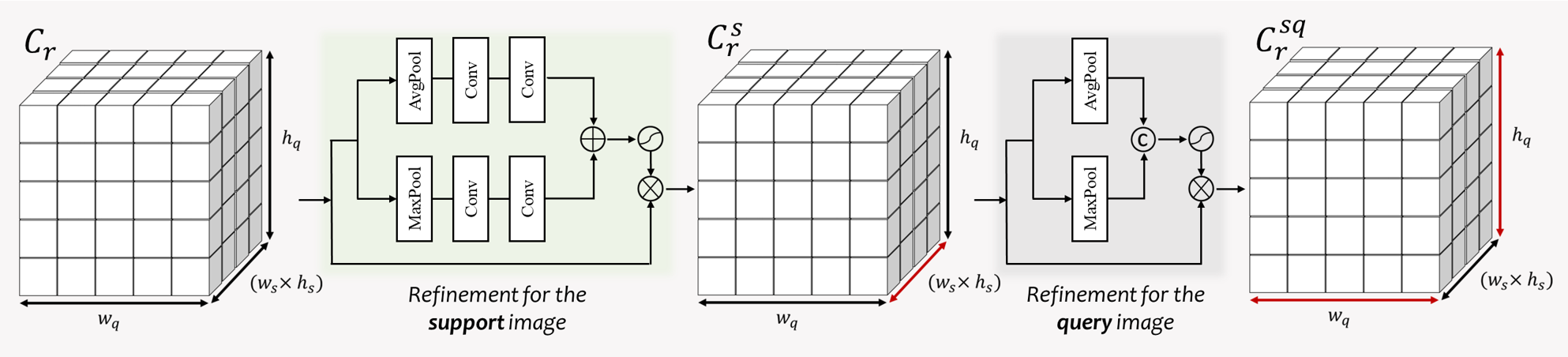}
	\end{center}
	\vspace{-0.5cm}
	\renewcommand{\figurename}{\textbf{Fig.}}
	\caption{Flowchart of dual spatial attention block. Given the correlation map $C_{r}$ from the correlation matching module, the first step is to attention to the dimension $d = (h_{s} \times w_{s})$ of the correlation map, which can be regarded as a refinement for the support image. The next step is to attention to the dimensions $w_{q}$ and $h_{q}$ of the correlation map, which can be regarded as a refinement for the query image. }
	\label{fig:KAB}
	\vspace{-0.5cm}
\end{figure*}
\subsubsection{Dual Spatial Attention Block}
When humans are guided by support images, during a glance, human will remember the most critical salient regions that contribute to visual spotting. Besides, when we need to find a character in a page of a book, blank areas or illustrations are ignored in the process of visual spotting. Thus, the spotting area is narrowed to concentrate on potential regions of interest within the query image, which means human will focus potential discriminative area.

\textbf{\textit{Step 1. }Refinement for the support image.} Given the correlation map $C_{r} \in R^{ h_{q} \times w_{q} \times d}$ by the correlation matching module, where $d=( h_{s} \times w_{s})$. The correlation map not only contains all information from the support and query images, but also captures the similarity between the two images. For correlation map $C_{r}$, each channel $d_{j}$ of $C_{r}$ represents the correlation of the $j$-th pixel on the feature map $\{f_{s}\}_{j}$ with $f_{q}$, we produce a 1d spatial-wise attention map to focus ``where" is meaningful localization in the support image, as shown in \cref{fig:KAB}. We first feed the obtained $C_{r}$ into the dual spatial attention block to find regions with key discriminative properties in the support image. Aggregating the spatial information of the correlation map $C_{r}$ by using average-pooling and max-pooling operations, generating two different spatial context descriptors $f^{avg}_{c}$ and $f^{max}_{c}$. After that, features pooled by each pooling layer are then passed through the support refinement network, which consists of two convolutions layer with kernel size 1. The support refinement network analyzes the correlation map $C_{r}$ to identify the most important regions in the support image by learning the relationships between different channels. Then the element-wise summation is used to merge the results which have:  
\begin{equation}
	\begin{split}
		M_{s} = &\sigma_{s}(W_{1}(W_{0}(AvgPool(C_{r})))\\
		&+W_{1}(W_{0}(MaxPool(C_{r})))),
	\end{split}
\end{equation}
where $W_{0} \in R^{d\times (d/\tau)}$, $W_{1} \in R^{(d/\tau) \times d}$ are parameters of convolutions, $\sigma_{s}$ represents the sigmoid activation function and $\tau$ is the reduction ratio. Then multiply $M_{s} \in R^{1\times 1 \times d}$ with the correlation map $C_{r}$, expressed as:
\begin{equation}
	\begin{split}
		C_{r}^{s} = M_{s} \otimes C_{r},
	\end{split}
\end{equation}
where $C_{r}^{s}$ captures the key points in spatial dimension of the support image and $\otimes$ denotes element-wise multiplication.

\textbf{\textit{Step 2.} Refinement for the query image.} Note that, at a particular position $(a,b)$ of $C_{r}^{s}$ contains the similarities between $f_{q}(a,b)$ and all the features of $f_{s}$. As illustrated in \cref{fig:KAB}, we also generate a 2d spatial attention map to focus on the positional informative key point for the query image. By using average-pooling and max-pooling operations along the channel dimension and concatenating the pooled features. Then a convolution layer is applied to generate the spatial attention map:
\begin{equation}
	\begin{split}
		M_{q} = &\sigma_{q}(Conv([AvgPool(C_{r}^{s});MaxPool(C_{r}^{s})])),
	\end{split}
\end{equation}
where $\sigma_{q}$ represents the sigmoid activation function, and $Conv$ represents a convolution operation with the filter size of $3 \times 3$. To obtain the final refined features, we multiply $M_{q} \in R^{h_{q} \times w_{q} \times 1}$ with $C_{r}^{s}$, which can be expressed briefly as:
\begin{equation}
	\begin{split}
		C_{r}^{sq} = M_{q} \otimes C_{r}^{s},
	\end{split}
\end{equation}
where $C_{r}^{sq}$ captures the key points in the spatial dimension of the query image and retains the key points focus of the support image.
\subsubsection{Geometric Matching Block}
For the obtained attention correlation map $C_{r}^{sq}$, the geometric matching block is applied to find the spatial correspondence between $I_{s}$ and $I_{q}$.
To achieve this, we optimize a transformation function $W_{\theta}:\mathbb{R}^{2} \rightarrow \mathbb{R}^{2}$, where $\theta$ denotes the alignment parameter. 
The spatial coordinates between $I_{q}$ and $I_{s}$ can be found by $(k’,l’) = W_{\theta}(k,l)$, where $(k’,l’)$ are the corresponding spatial coordinates of $(k,l)$ in $f_{q}$.
Similar to \citep{MATCH01}, the geometric matching block consists of two consecutive convolutional layers without padding and stride equal to 1, with batch normalization and ReLU connected between the two convolutional layers, and finally a fully-connected layer to regress the alignment parameter $\theta$. 
The goal of the GM block is to find the best parameters $\hat{\theta}$:
\begin{equation}
	\hat{\theta} = \underset{\theta}{\arg \max} \sum_{(k,l)} \phi(f^{q}_{W_{\theta}(k,l)} , f^{s}_{k,l}),
\end{equation}
where $\underset{(k,l)}{\sum} \phi(f^{q}_{W_{\theta}(k,l)}, f^{s}_{k,l})$ means the summation of the feature similarities between the support feature and the regions corresponding to those on the query image obtained through the 2D geometric transformation. After that, the obtained transformations are input to the grid sampler to generate a grid of points, which is aligned with the support image at each location of the query image. 

Note that, the process of location is already done implicitly in the alignment process, by extracting the position frames based on the maximum and minimum values of the obtained grid tensor, by simply outputting a square tight bounding box based on the transformed grid points. The spotting score can be obtained by computing $\underset{(k,l)}{\sum} \phi(f^{q}_{W_{\theta}(k,l)}, f^{s}_{k,l})$, \textit{i.e.}, the similarity score of the local region on the query image feature $f^{q}_{W_{\theta}(k,l)}$ to the input support image feature $f^{s}_{k,l}$.

\subsection{Loss Function}
The loss function needs to construct the training objective from both localization and spotting \citep{LOSS01, OS02, OS03}. 

\textbf{Localization Loss.} In this study, the smooth $L_{1}$ loss is used for localization: 
\begin{equation}
	l_{loc}(v,t) =
	\begin{cases}
		\underset{p} {\sum} \frac{1}{2}(v_{p}-t_{p})^{2},           & \text{if $\vert v_{p} – t_{p} \vert < 1$}.\\
		\underset{p} {\sum} \vert v_{p} – t_{p} \vert – \frac{1}{2},& \text{otherwise.}
	\end{cases}
\end{equation}
where $v = (v_{x},v_{y},v_{w},v_{h})$ denotes the frame coordinates of GT and $t = (t_{x},t_{y},t_{w},t_{h})$ denotes the frame coordinates obtained by prediction. 

\textbf{Spotting Loss.}  Our task has an inherent difficulty to balance positive and negative samples. Ideally, the distance metric of positive examples is infinite, while the distance metric of negative examples should be close to 0. To separate positive and negative samples, the distance metric of positive samples should be greater than margin $m_{pos}$, while the distance metric of negative samples should be less than margin $m_{neg}$. Therefore, the hinge-embedding loss with margins can be written as follows:
\begin{equation}
	\begin{split}
		l_{pos}^{i}= max(m_{pos}-s_{i},0),
	\end{split}
\end{equation}
\begin{equation}
	\begin{split}
		l_{neg}^{i}= max(s_{i}-m_{neg},0),
	\end{split}
\end{equation}
where $m_{pos}$ and $m_{neg}$ are the positive and negative margins, and $s\in \left[ -1 ,1 \right]$ is the spotting score. To separate positive and negative examples, the contrastive loss \citep{Contrastive} is being considered:
\begin{equation}
	\begin{split}
		L_{c} &= \underset{i} {\sum} (l_{pos}^{i} + l_{neg}^{i}).
	\end{split}
\end{equation}
To acquire discriminative embeddings through contrastive learning, \citet{LOSS02} introduced the ranked list (RL) loss which can adaptively mine negative samples by providing a weighting parameter. The RL loss can be formulated as follows: 
\begin{equation}\label{RL}
	\begin{split}
		L_{rl} &= \underset{i}{\sum} (l_{pos}^{i} + w_{i} l_{neg}^{i}),
	\end{split}
\end{equation}
where $w_{i}$ is defined as $w_{i}=exp(T\cdot(s_{i}-m_{neg})), s_{i}-m_{neg}>0$, and $T$ is the temperature parameter which controls the degree of weighting negative examples. However this loss function can not deal with the challenging examples. 

\begin{figure}[!t]
	\begin{center}
		\includegraphics[clip,width=8.5cm]{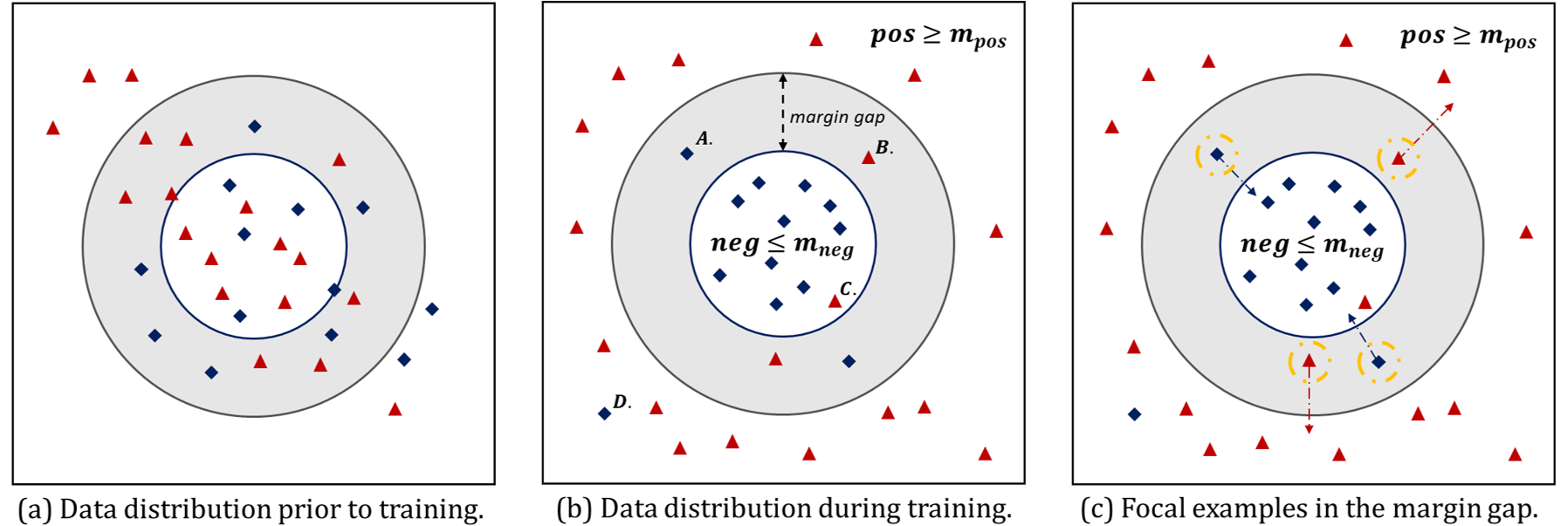}
	\end{center}
	\vspace{-0.5cm}
	\renewcommand{\figurename}{\textbf{Fig.}}
	\caption{Data distribution analysis. (a) Prior to training, the data may be distributed in a complex and nonlinear fashion, with no clear separation between positive and negative examples. (b) During training, the model may be able to identify some patterns and achieve a degree of separation between positive and negative examples. Some examples may still remain in the margin gap, where the decision boundary is unclear and further optimization is needed to achieve greater separation. (c) To achieve optimal performance, we expect the values of positive examples to be greater than the margin $m_{pos}$, while the values of negative examples should be less than the margin $m_{neg}$.}
	\label{fig:loss}
	\vspace{-0.5cm}
\end{figure}

The RL loss is designed to assign different weights to negative examples and measure the distance between $s_{i}$ and $m_{neg}$, thereby alleviating the issue of imbalanced examples. \cref{fig:loss}(a) depicts the data distribution before training. During training  (\cref{fig:loss}(b)), examples C and D are expected to incur a large loss from the RL loss (Eq. (\ref{RL})). However, the challenging examples A and B within the margin gap will receive a relatively small loss,  which is suboptimal for achieving the desired separation between positive and negative examples. To address this issue, it is desirable for the values of positive examples to exceed the margin threshold $m_{pos}$, whereas the values of negative examples should fall below the margin $m_{neg}$. Accordingly, we aim to focus on the challenging examples in the margin gap and give them more attention to achieve a clear separation between positive and negative examples. For the example in the margin gap, we set:
\begin{equation}
	g_{i}=
	\left\{
	\begin{array}{rl}
		m_{pos}, & s_{i} \ge m_{pos} \\
		s_{i},   & m_{neg} < s_{i} < m_{pos}\\
		m_{neg}, & s_{i} \le m_{neg} 
	\end{array}
	\right.
\end{equation}

Then the improved hinge-embedding loss with margins can be formulated as follows:
\begin{equation}
	\begin{split}
		l_{pos'}^{i}= l_{pos}^{i} - log(\frac{g_{i}}{m_{pos}}),
	\end{split}
\end{equation}
\begin{equation}
	\begin{split}
		l_{neg'}^{i}= w_{i}l_{neg}^{i} - log(\frac{m_{pos}+m_{neg}-g_{i}}{m_{pos}}),
	\end{split}
\end{equation}
where $w_{i}$ is also defined as $w_{i}=exp(T\cdot(s_{i}-m_{neg})), s_{i}-m_{neg}>0$. When the $s_{i}$ falls in the margin gap, both positive and negative examples receive extra loss, which help to further separate them. For the $-log(g_{i}/m_{pos})$, we want the value of positive examples to exceed the margin threshold $m_{pos}$. If the spotting score is greater than $m_{pos}$, it meets the optimization goal and no further optimization is necessary. Similarly, for the $-log((m_{pos}+m_{neg}-g_{i})/(m_{pos}))$, we expect the value of negative examples to fall below the margin threshold $m_{neg}$. 

The new loss function which improved RL loss which called torus loss can be formulated as follows:
\begin{equation}
	\begin{split}
		L_{torus} &= \underset{i}{\sum} (l_{pos'}^{i} + l_{neg'}^{i}).
	\end{split}
\end{equation}

The torus loss not only addresses the issue of challenging examples in the margin gap, but also helps to mitigate the impact of example imbalance. Therefore, we can use this function as the spotting loss.

\textbf{Training Objective.} Finally, the total loss of our model can be expressed as:
\begin{equation}
	L = \lambda L_{loc} + L_{torus},
\end{equation}
where $L_{loc} = \underset{i}{\sum}l_{loc}(v_{i},t_{i})$ and $L_{torus}=\underset{i}{\sum} (l_{pos'}^{i} + l_{neg'}^{i})$. $\lambda$ is a hyperparameter that balances the importance of the two loss terms.

\section{Experiments}
\subsection{Datasets}
To thoroughly evaluate the efficacy of our proposed method and contribute meaningfully to the research community, we have meticulously curated existing public datasets and created a new dataset. 

\textbf{Dongba Hieroglyphics dataset (DBH)} is a new dataset featuring Dongba characters, an ancient script created by the Naxi minority's ancestors in China. These pictographs hold historical and literary value and have been recognized as "Memory of the World" by UNESCO. As the total number of distinct Dongba characters is unknown, text spotting in manuscripts is an open-set problem, where spotting novel characters helps decipher historical texts. We collected data from Dongba sutras, annotating and summarizing it into a dataset of 3,633 bounding boxes across 253 categories. To accommodate one-shot tasks, experts hand-wrote an additional 253 characters, using them as support images. The dataset is available for download\footnote{\url{https://github.com/infinite-hwb/VGTS/tree/master/DATA}}.

\textbf{Egyptian Hieroglyph Dataset (EGH)} \citep{EGY} built from hieroglyphs in 10 different pictures from the book 'The Pyramid of Unas', comprises 171 classes. For this dataset, 4 images are selected for training, while 6 images are used for testing. It notably includes 68 novel categories, not present during the training process.

\textbf{VML-HD dataset (VML)} \citep{VML} is a valuable resource for researchers working with historical Arabic documents. It comprises 94 images, with a total of 1,770 unique character categories, of which 150 are randomly selected as novel categories. The training set includes 65 images, and the remaining 29 images are used for testing. 

\textbf{Tripitaka Koreana in Han dataset (TKH)} \citep{TKH} consists of scanned images from Chinese antiquarian books. It contains a total of 999 images, with 1,492 different character categories. For this dataset, 171 images are chosen for training, and 828 images are reserved for testing.  

\textbf{Notary Charters dataset (NC)} \citep{NOTARY} is a collection of manuscripts from classical and medieval eras, consisting of 388 different categories of notary charters. Among these, 311 categories appear only once, and are thus selected for training. The remaining images are reserved for testing, with 77 categories chosen as novel categories.  
\subsection{Implementation details}
We employ a data augmentation strategy that involves randomly cropping patches of size 800$\times$800 to all query images. This approach not only prevents over-fitting, but also allows the model to learn from input query images containing multiple characters and categories. Besides, we iteratively adjusted it based on the results obtained from the validation set and selected the model that achieved the best performance. For the training objective function, the $m_{pos}=0.6$ and $m_{neg}=0.5$, and the hyperparameter $\lambda$ is set to 0.2. 
Our method is based on PyTorch 1.9 and trained using the Adam optimizer \citep{ADAM} with a learning rate of 1e-4 and a beta1 value of 0.9 on one Nvidia GeForce RTX 3090. We consider that the top three blocks of the pre-trained ResNet-50 network \citep{RES50}, used on ImageNet \citep{IMAGENET}, are used as the feature extractor. We apply an affine transformation with a 6-degree-of-freedom linear transformation for the geometric matching. To evaluate the performance of our model, we use the standard Pascal VOC metric \citep{VOC}, following the mainstream setting of using mean average precision (mAP), recall and F1 score as the evaluation metrics. We use an intersection-over-union (IoU) threshold of 0.5 for mAP evaluation. All datasets and code will be publicly available\footnote{\url{https://github.com/infinite-hwb/VGTS}}. To mitigate memory strain, the TKH dataset's test set is divided into Set 1 and Set 2, comprising 390 and 438 test images, respectively. During evaluation, the support image is resized while maintaining its aspect ratio for all comparative methods. A five-level pyramid (0.4, 0.6, 0.8, 1.0, and 1.2 times the dataset scale) is employed for the query image. The DBH dataset has a scale of 2500, the VML dataset has a scale of 4000, the TKH dataset has a scale of 5500, and the NC dataset has a scale of 4000.

\subsection{Results}
\subsubsection{Comparison Baseline}
Initially, it is hypothesized that certain one-shot object detection methods could also address the task of one-shot text spotting. To test this hypothesis, we retrained some popular few-shot object detection methods \citep{OS03} using the four datasets previously summarized. Unfortunately, the experimental performance fell short, struggling to surpass 8$\%$ mAP. This is mainly due to the fact that historical manuscripts usually comprise a large number of characters, frequently exceeding 100 different categories. These characters possess similar glyphs or significantly vary in scale, thereby presenting significant challenges for one-shot text spotting. Although we attempted to use the popular object detector method Faster RCNN \citep{RPN}, the paucity of training data made it difficult to achieve meaningful results. Therefore, we opted to use OS2D \citep{OS02}, a dense correlation matching feature-based method for one-shot object detection, as our baseline. This approach trains the transform network from a model pre-trained with weak supervision. As there has been no prior research on one-shot text spotting, we adapted OS2D \citep{OS02} to suit our historical manuscripts by fine-tuning the method's parameters and using the same data augmentation scheme as our proposed method. Although CoAE's intended purpose is to detect objects in natural scenes, it has been included as one of the comparison methods for its efficacy in one-shot object detection using the RPN approach in recent years. To make a fair comparison, we have also added multi-scale pyramids to the training and testing of CoAE. Furthermore, we have experimented with DSW \citep{DSW}, a classic unsupervised approach for text spotting based on the ``sliding window" technique, and reproduced its main idea by directly mapping the feature map of the support image as a convolution on the query image after extracting the features using a Siamese network with specifications identical to our method.

\begin{table}[!t]
	\vspace{-0.5cm}
	\renewcommand{\arraystretch}{1.1}
    \setlength{\tabcolsep}{5pt}
	\caption{\\ Quantitative evaluation on DBH, EGH, VML, TKH and NC on the ``Novel" class. The best results are highlighted in \textbf{bold}.}
	\begin{tabular*}{\hsize}{@{}@{\extracolsep{\fill}}llcccc@{}}
		\hline
		\rule{0pt}{3pt} Dataset               & Method        & Trained & mAP  & Recall & F1\rule{0pt}{3pt}\\
		\hline
		\rule{0pt}{3pt} \multirow{4}*{DBH}    
		       &DSW           & No      & 51.70& 59.64 & 55.39 \rule{0pt}{3pt}\\ 
		~      &CoAE          & Yes     & 62.64& 60.36 & 61.48 \rule{0pt}{3pt}\\
		~      &OS2D          & Yes     & 98.08& 99.29 & 98.68 \rule{0pt}{3pt}\\  
		~      &VGTS(Ours)    & Yes     &\textbf{99.85} &\textbf{100.0} & \textbf{99.92} \rule{0pt}{3pt}\\
		\hline		
		\rule{0pt}{3pt} \multirow{4}*{EGH}    
		       &DSW           & No      & 51.88& 72.86 & 60.61 \rule{0pt}{3pt}\\
		~      &CoAE          & Yes     & 12.59& 55.06 & 20.50 \rule{0pt}{3pt}\\   
		~      &OS2D          & Yes     & 78.63& 91.43 & 84.55 \rule{0pt}{3pt}\\  
		~      &VGTS(Ours)    & Yes     &\textbf{83.86} &\textbf{97.62} & \textbf{90.22} \rule{0pt}{3pt}\\
		\hline		
		\rule{0pt}{3pt} \multirow{4}*{VML}    
		       &DSW           & No      & 62.77& 74.15 & 67.99 \rule{0pt}{3pt}\\
		~      &CoAE          & Yes     & 39.07& 60.17 & 47.38 \rule{0pt}{3pt}\\   
		~      &OS2D          & Yes     & 94.39& 98.31 & 96.31 \rule{0pt}{3pt}\\  
		~      &VGTS(Ours)    & Yes     &\textbf{100.0} &\textbf{100.0} & \textbf{100.0} \rule{0pt}{3pt}\\
		\hline		
		\rule{0pt}{3pt} \multirow{4}*{TKH Set 1}  
		       &DSW           & No      & 42.63& 92.02 & 58.27 \rule{0pt}{3pt}\\
		~      &CoAE          & Yes     & 36.74& 47.30 & 41.36 \rule{0pt}{3pt}\\      
		~      &OS2D          & Yes     & 88.05& 99.60 & 93.47 \rule{0pt}{3pt}\\  
		~      &VGTS(Ours)    & Yes     & \textbf{90.20}&\textbf{99.65} & \textbf{94.69} \rule{0pt}{3pt}\\
		\hline
		\rule{0pt}{3pt} \multirow{4}*{TKH Set 2}  
		       &DSW           & No      & 37.95& 88.68 & 53.15 \rule{0pt}{3pt}\\
		~      &CoAE          & Yes     & 35.78& 45.73 & 40.15 \rule{0pt}{3pt}\\      
		~      &OS2D          & Yes     & 86.20& 98.35 & 80.55 \rule{0pt}{3pt}\\  
		~      &VGTS(Ours)    & Yes     & \textbf{88.06}&\textbf{99.52} & \textbf{93.44} \rule{0pt}{3pt}\\
		\hline
		\rule{0pt}{3pt} \multirow{4}*{NC}         
		       &DSW           & No      & 59.60& 66.06 & 62.66 \rule{0pt}{3pt}\\
		~      &CoAE          & Yes     & 85.82& 93.58 & 89.53 \rule{0pt}{3pt}\\      
		~      &OS2D          & Yes     & 76.15& 77.92 & 77.02 \rule{0pt}{3pt}\\  
		~      &VGTS(Ours)    & Yes     & \textbf{92.14}&\textbf{96.79} & \textbf{94.41} \rule{0pt}{3pt}\\
		\hline
	\end{tabular*}
	\label{tab:TESTThree}
	\vspace{-0.5cm}
\end{table}
\subsubsection{Statistical Results and Analysis}
The statistical results for the \textit{Novel} class across various datasets are presented in \cref{tab:TESTThree}. The \textit{Novel} class denotes an unseen category absent during training. In contrast, the statistical outcomes for the \textit{Base} class across different datasets are displayed in \cref{tab:TESTFour}, with the \textit{Base} class signifying a seen category present during training. 

Experimental results reveal that our method attains the highest recall score across all datasets, suggesting a low propensity for missed text spotting occurrences. Additionally, our proposed method surpasses the other methods in mAP and F1 scores. It is pertinent to acknowledge that the sliding window method may be deemed an unsupervised technique, and the accurate aspect ratio of support image feature maps is provided for this approach. Nevertheless, the performance of the unsupervised sliding window method is inferior to supervised learning methods, such as OS2D and VGTS. Moreover, the unsupervised DSW method occasionally outperforms the supervised CoAE method, likely due to inadequate data for comprehensive CoAE learning. Regarding data distribution, the DBH, EGH and NC datasets exhibit a more dispersed inter-class character distribution in manuscript images, whereas the VML and TKH datasets display a more compact inter-class distribution and smaller character size, which CoAE struggles to accommodate. With the EGH, VML and TKH datasets containing numerous similar characters, our VGTS method outperforms the baseline method. We also observed that performance on novel categories occasionally exceeds that of base categories, a phenomenon evident across multiple methods. This observation may be attributed to the relatively smaller data volume for novel categories, which can significantly impact their precision and recall rates during evaluation. Specifically, when a method accurately spots limited samples of novel categories, precision may increase, resulting in higher Average Precision (AP) for those categories.
\begin{table}[!t]
	\vspace{-0.5cm}
	\renewcommand{\arraystretch}{1.1}
	\setlength{\tabcolsep}{5pt}
	\caption{\\ Quantitative evaluation on DBH, VML and TKH on the ``Base" class. Note that, the test set of the NC does not contain the base categories, and hence, we only evaluate the other three datasets. The best results are highlighted in \textbf{bold}.}
	\begin{tabular*}{\hsize}{@{}@{\extracolsep{\fill}}llcccc@{}}
		\hline
		\rule{0pt}{3pt} Dataset               & Method        & Trained & mAP  & Recall &F1 \rule{0pt}{3pt}\\
		\hline
		\rule{0pt}{3pt} \multirow{4}*{DBH}    
		       &DSW           & No   &39.08 &35.76 &37.35 \rule{0pt}{3pt}\\ 
		~      &CoAE          & Yes  &72.55 &93.55 &81.72 \rule{0pt}{3pt}\\
		~      &OS2D          & Yes  &89.22 &94.76 &91.91 \rule{0pt}{3pt}\\  
		~      &VGTS(Ours)    & Yes  &\textbf{91.74}&\textbf{95.17} & \textbf{93.42} \rule{0pt}{3pt}\\
		\hline		
		\rule{0pt}{3pt} \multirow{4}*{EGY}    
		       &DSW           & No   & 18.44& 39.36 & 25.11 \rule{0pt}{3pt}\\
		~      &CoAE          & Yes  & 13.43& 50.48 & 21.22 \rule{0pt}{3pt}\\   
		~      &OS2D          & Yes  & 52.92& 77.00 & 62.73 \rule{0pt}{3pt}\\  
		~      &VGTS(Ours)    & Yes  &\textbf{76.13} &\textbf{97.19} & \textbf{85.38} \rule{0pt}{3pt}\\
		\hline		
		\rule{0pt}{3pt} \multirow{4}*{VML}    
		       &DSW           &No    & 20.70 &46.83 &28.71 \rule{0pt}{3pt}\\
		~      &CoAE          &Yes   & 27.30 &53.07 &36.05 \rule{0pt}{3pt}\\   
		~      &OS2D          &Yes   & 55.31 &52.45 &53.84 \rule{0pt}{3pt}\\  
		~      &VGTS(Ours)    &Yes   &\textbf{66.59}&\textbf{70.57}&\textbf{68.52} \rule{0pt}{3pt}\\
		\hline		
		\rule{0pt}{3pt} \multirow{4}*{TKH Set 1}  
		       &DSW           &No    &41.70&83.71&55.67 \rule{0pt}{3pt}\\
		~      &CoAE          &Yes   &58.81&85.92&69.83 \rule{0pt}{3pt}\\      
		~      &OS2D          &Yes   &89.31&95.72&92.40  \rule{0pt}{3pt}\\  
		~      &VGTS(Ours)    &Yes   &\textbf{90.51}&\textbf{95.78}&\textbf{93.07} \rule{0pt}{3pt}\\
		\hline
		\rule{0pt}{3pt} \multirow{4}*{TKH Set 2}  
		       &DSW           &No    &36.22&82.03 &50.25\rule{0pt}{3pt}\\
		~      &CoAE          &Yes   &54.67&81.74&65.52 \rule{0pt}{3pt}\\      
		~      &OS2D          &Yes   &85.53&95.96&90.45 \rule{0pt}{3pt}\\  
		~      &VGTS(Ours)    &Yes   &\textbf{87.50}&\textbf{96.26}&\textbf{91.67} \rule{0pt}{3pt}\\
		\hline
	\end{tabular*}
	\label{tab:TESTFour}
\end{table}

\begin{table}[!t]
	\centering
	\vspace{-0.3cm}
	\renewcommand{\arraystretch}{1.1}
	\caption{\\ Performance evaluation for similar characters, ``num” is the number of occurrences for each character.}
	\begin{subtable}{.5\linewidth}
		\centering
		\caption{``\includegraphics[scale=0.015]{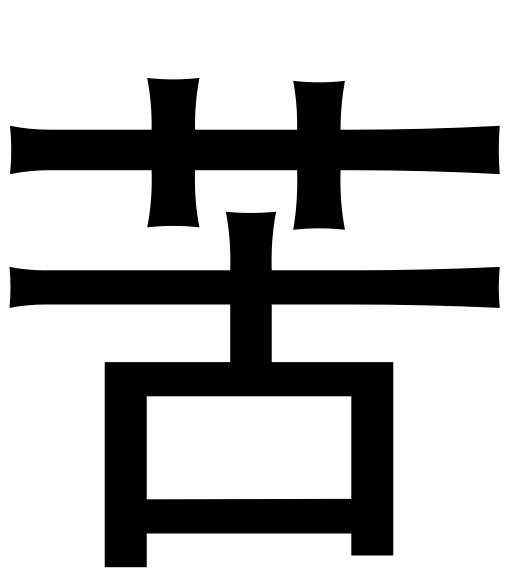}"(\textit{Novel}) and ``\includegraphics[scale=0.015]{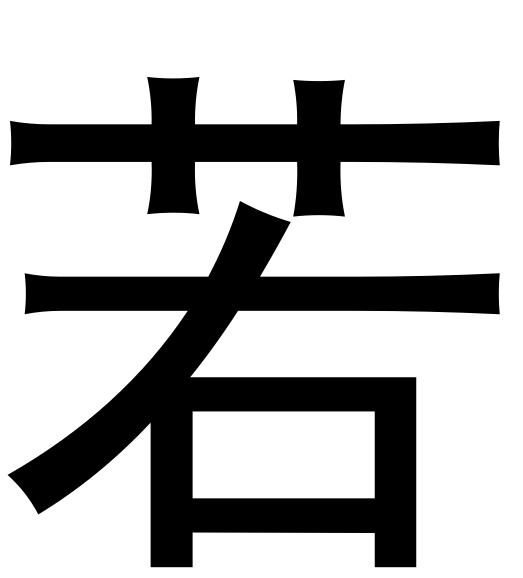}"(\textit{Base}).}
		\begin{tabular}{l|l|l}
			\hline
			Class & AP$_{50}$ & \# num. \\
			\hline
			``\includegraphics[scale=0.015]{1012.jpg}" & 94.47 &421\\
			``\includegraphics[scale=0.015]{105.jpg}"  & 95.04 &11,017\\
			\hline
		\end{tabular}
	\end{subtable}%
	\begin{subtable}{.5\linewidth}
		\centering
		\caption{``\includegraphics[scale=0.015]{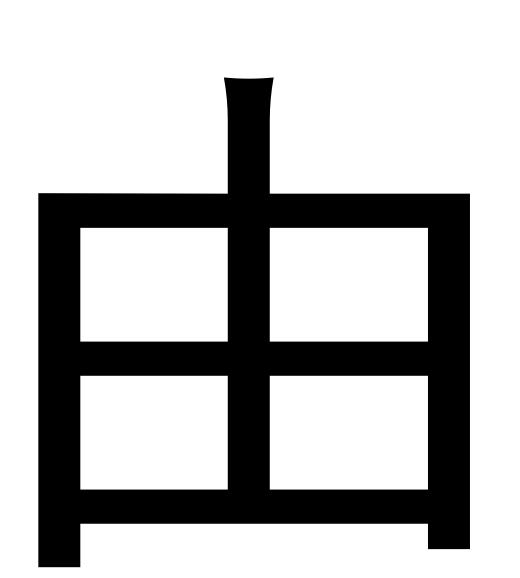}"(\textit{Novel}) and ``\includegraphics[scale=0.015]{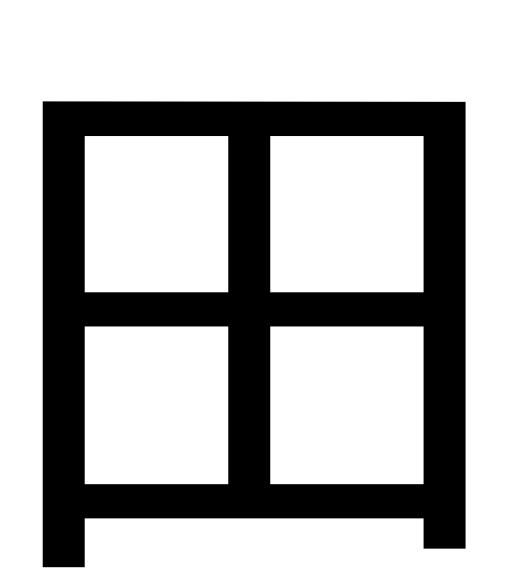}"(\textit{Base}).}
		\begin{tabular}{l|l|l}
			\hline
			Class & AP$_{50}$ & \# num. \\
			\hline
			``\includegraphics[scale=0.015]{429.jpg}" & 98.50 &186\\
			``\includegraphics[scale=0.015]{214.jpg}" & 100.0 &3\\
			\hline
		\end{tabular}
	\end{subtable}
	\begin{subtable}{.5\linewidth}
		\centering
		\caption{``\includegraphics[scale=0.015]{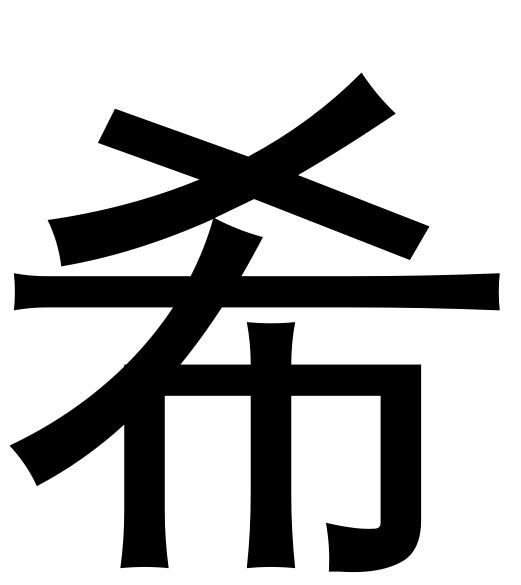}"(\textit{Novel}) and ``\includegraphics[scale=0.015]{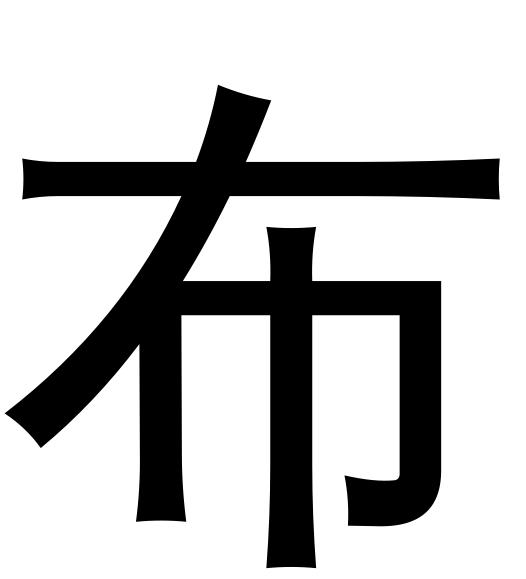}"(\textit{Base}).}
		\begin{tabular}{l|l|l}
			\hline
			Class & AP$_{50}$ & \# num.\\
			\hline
			``\includegraphics[scale=0.015]{627.jpg}"  & 62.5 &2\\
			``\includegraphics[scale=0.015]{475.jpg}"  & 99.41 &237\\
			\hline
		\end{tabular}
	\end{subtable}%
	\begin{subtable}{.5\linewidth}
		\centering
		\caption{``\includegraphics[scale=0.015]{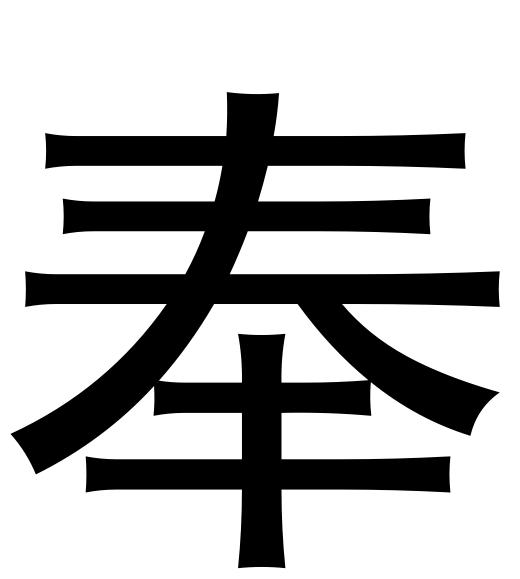}"(\textit{Novel}) and ``\includegraphics[scale=0.015]{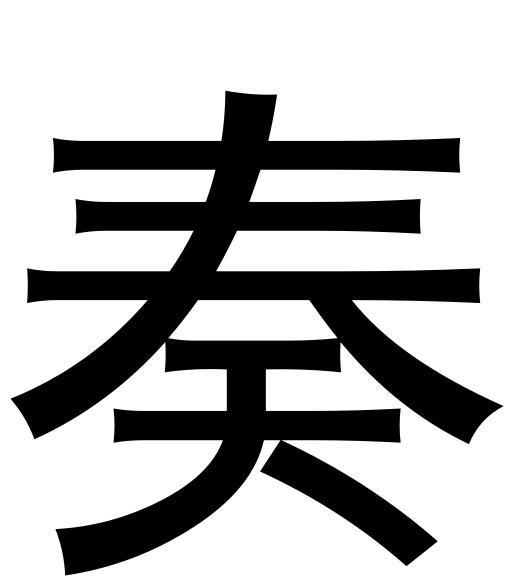}"(\textit{Base}).}
		\begin{tabular}{l|l|l}
			\hline
			Class & AP$_{50}$ & \# num.\\
			\hline
			``\includegraphics[scale=0.015]{655.jpg}"  & 85.72 &72\\
			``\includegraphics[scale=0.015]{1407.jpg}" & 93.33 &2\\
			\hline
		\end{tabular}
	\end{subtable}
	\label{tab:NUM}
	\vspace{-0.5cm}
\end{table}

During training, VGTS learns to match potential query image regions with the support character instance rather than accumulating class-specific knowledge. This learning strategy facilitates effective performance with limited training samples, crucial in the context of historical manuscript data featuring small, sparse datasets. By circumventing the need for an extensive training sample collection, the proposed method distinguishes itself from other one-shot methods that require larger sample volumes to achieve similar results. 

\textbf{Similar Characters Challenge.} 
Misidentification can occur due to the structural similarity between characters, particularly when characters in `Novel' categories bear a high resemblance to those in `Base' categories. To assess this, we specifically selected `Base' and `Novel' categories, which consist of morphologically similar characters from the TKH dataset. As indicated in \cref{tab:NUM}, VGTS maintains high-performance metrics.

\begin{figure}[!t]
	\begin{center}
		\includegraphics[clip,width=8.5cm]{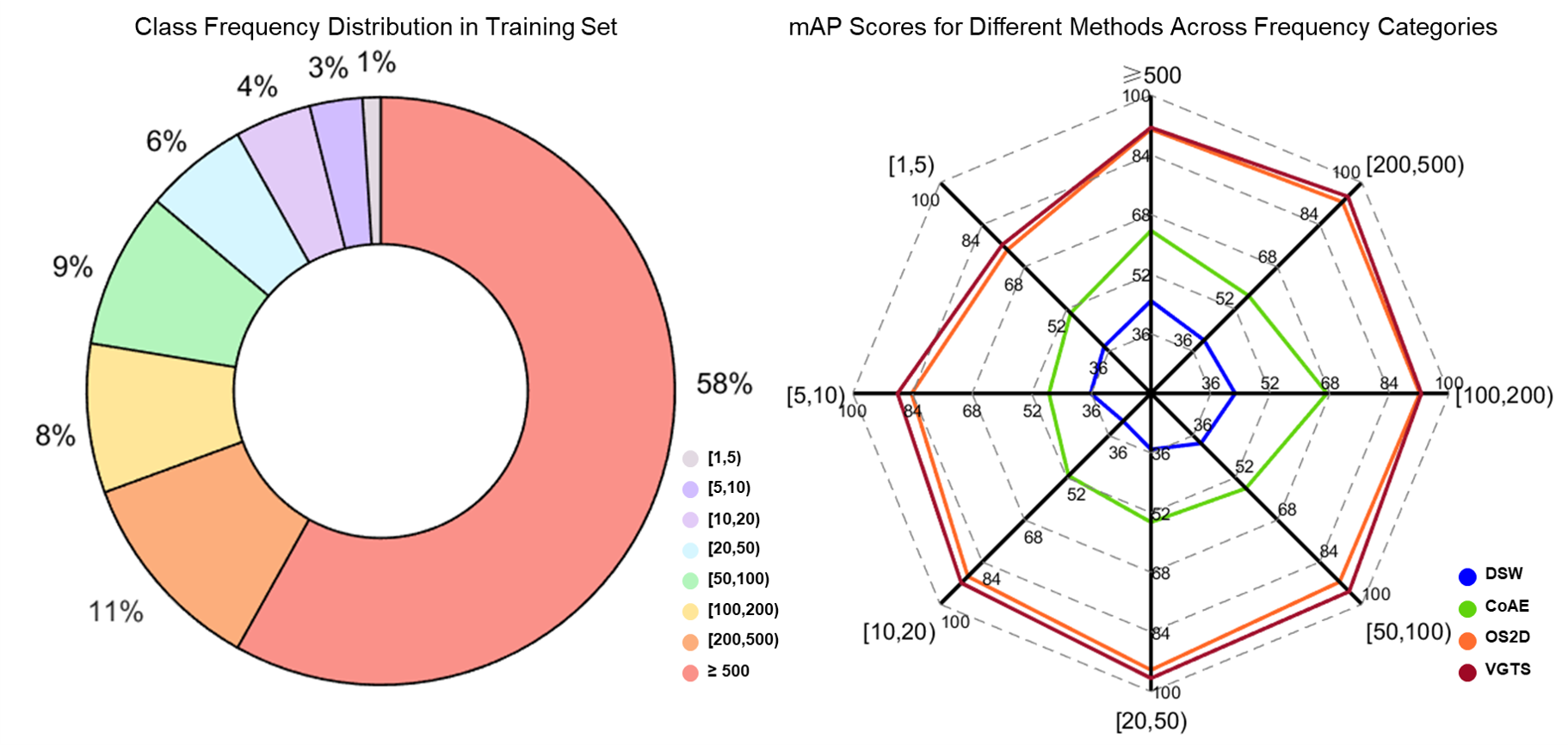}
	\end{center}
	\vspace{-0.5cm}
	\renewcommand{\figurename}{\textbf{Fig.}}
	\caption{Character frequency distribution in the TKH dataset and comparative mAP scores of different methods.}
	\label{fig:long}
	\vspace{-0.5cm}
\end{figure}

\textbf{Long-tailed Distribution Challenge.}
A challenge in text spotting in historical manuscript manuscripts is their long-tail distribution, with some characters appearing only once. Data from the TKH training set indicates that over 58\% of characters appear less than five times. This poses significant difficulties in training models, especially in capturing features of low-frequency characters. \cref{fig:long} compares the effectiveness of various methods in addressing this long-tail distribution issue. OS2D outperforms CoAE in high-frequency categories, while VGTS, our category-agnostic method, demonstrates higher mAP across all frequency categories, particularly in low-frequency ones such as `[1,5)' and `[5,10)'. This capability of VGTS is crucial for handling long-tail distributions, indicating its effectiveness in processing rare categories.

\begin{figure}[!h]
	\begin{center}
		\includegraphics[clip,width=8.5cm]{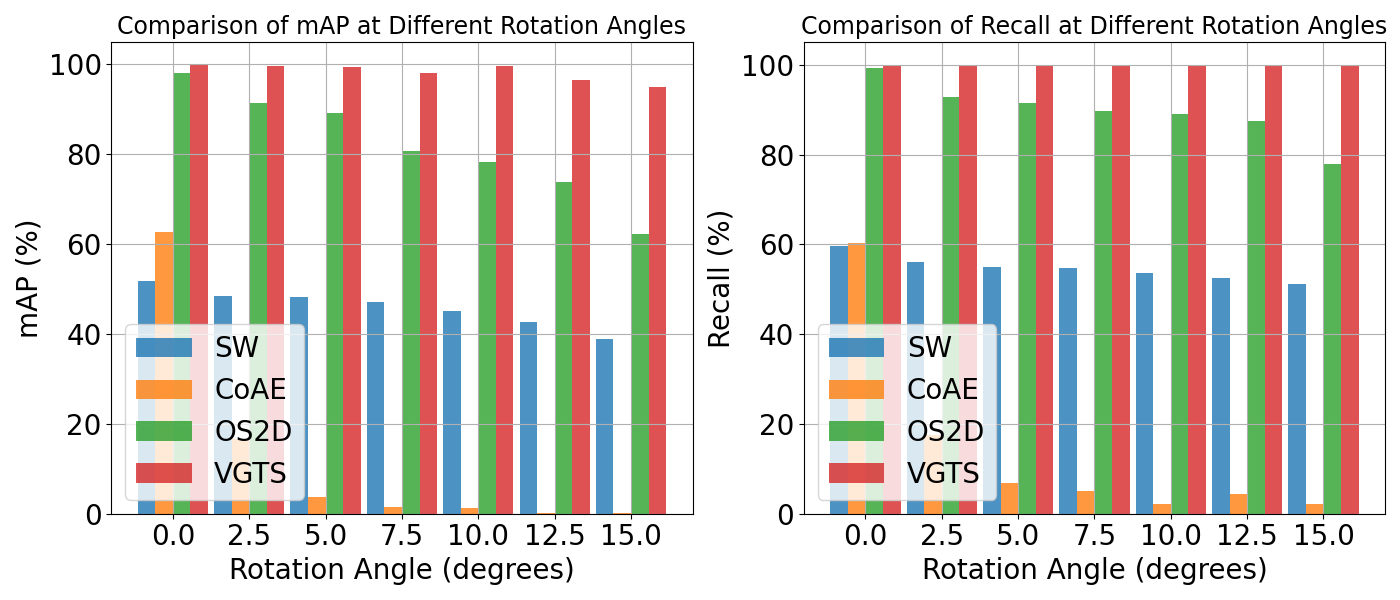}
	\end{center}
	\vspace{-0.5cm}
	\renewcommand{\figurename}{\textbf{Fig.}}
	\caption{Quantitative evaluation on DBH Novel with different angles of rotation on query images.}
	\label{fig:rotate}
	\vspace{-0.5cm}
\end{figure}

\textbf{Rotational Disturbance.} 
Our proposed method demonstrates exceptional robustness against minor rotational variations in test images, eliminating the need for rotation augmentations during training. We subjected test images to rotational disturbances ranging from 0$^{\circ}$ to 15$^{\circ}$ at 2.5$^{\circ}$ intervals. As depicted in \cref{fig:rotate}, our method surpasses the comparative methods in managing rotated test images, experiencing only minimal performance degradation as the rotation angle increases. Conversely, the RPN-based model, CoAE, exhibits a rapid decline in performance when confronted with rotational disturbances. The OS2D model's performance reveals a gradual degradation as the query image's rotation angle increases. The unsupervised sliding window-based method appears to be minimally impacted by rotational disturbances.
\subsubsection{Qualitative Results and Analysis}
\begin{figure}[!h]
	\begin{center}
		\includegraphics[clip,width=8.5cm]{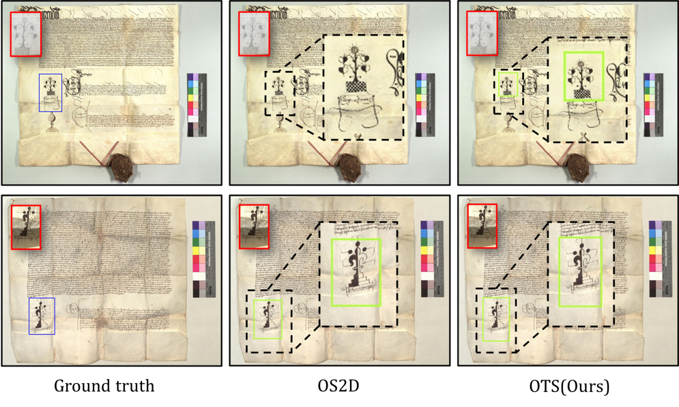}
	\end{center}
	\vspace{-0.5cm}
	\renewcommand{\figurename}{\textbf{Fig.}}
	\caption{Example results on NC dataset. The leftmost column shows the ground truth, and the red box is the support image. All prediction boxes are shown in green.}
	\label{fig:NV}
	\vspace{-0.3cm}
\end{figure}

\textbf{Spotting Performance Under Varied Image Conditions.} 
The support image quality substantially influences the localization and spotting performance. To demonstrate our method's robustness under varying image quality conditions, we present visual results with low-quality support images in \cref{fig:NV}. Remarkably, our method consistently delivers outstanding performance, even when the support image is blurred, as illustrated in the first row of \cref{fig:NV}. Both OS2D and VGTS exhibit accurate character spotting when the support image is affected by illumination, as shown in the second row of \cref{fig:NV}.

\begin{figure}[!h]
	\begin{center}
		\includegraphics[clip,width=8.5cm]{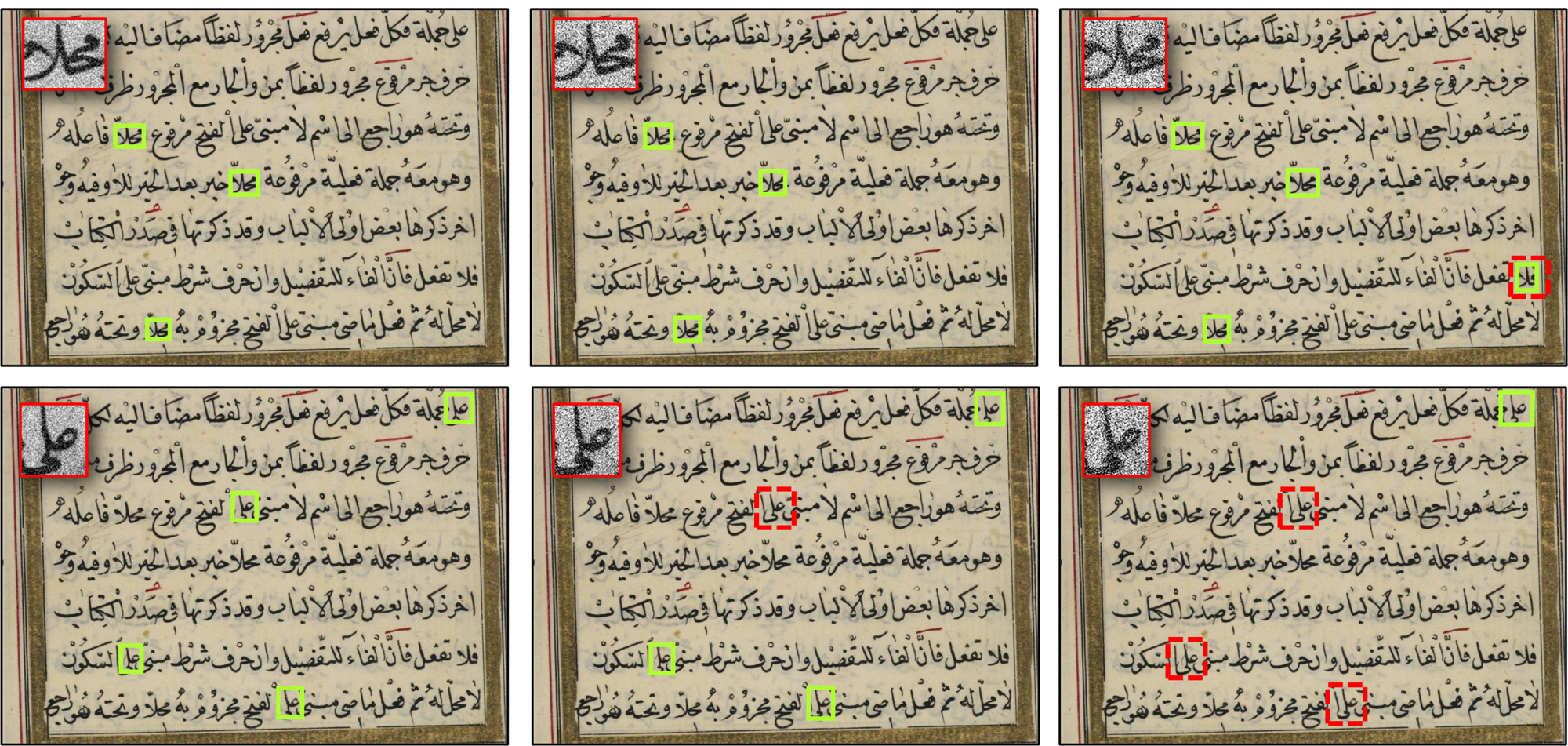}
	\end{center}
	\setlength{\abovecaptionskip}{0.1cm}
	\vspace{-0.3cm}
	\caption{Exemplary results on the VML dataset show the visual impact of different levels of Gaussian noise.}
	\label{fig:VMLV}
	\vspace{-0.3cm}
\end{figure}

We introduce Gaussian noise to the source images to simulate varying degrees of image degradation. The noise is added at variances corresponding to 15\%, 25\%, and 35\% of the full intensity scale, respectively. A qualitative analysis is illustrated in \cref{fig:VMLV}, with error detections highlighted by red dotted boxes. It is observed from \cref{fig:VMLV} that the model maintains robust performance under low to moderate levels of Gaussian noise. However, when the noise level reaches 25\%, the model begins to falter, resulting in instances of both missed and false spotting. We provide a demo video\footnote{\url{https://youtu.be/8GRDOxCDMjw}} \footnote{\url{https://www.bilibili.com/video/BV12x4y1K7MX/}} that effectively demonstrates the effectiveness of our proposed method, comprising several brief clips. 

\begin{figure}[!h]
	\begin{center}
		\includegraphics[clip,width=8.5cm]{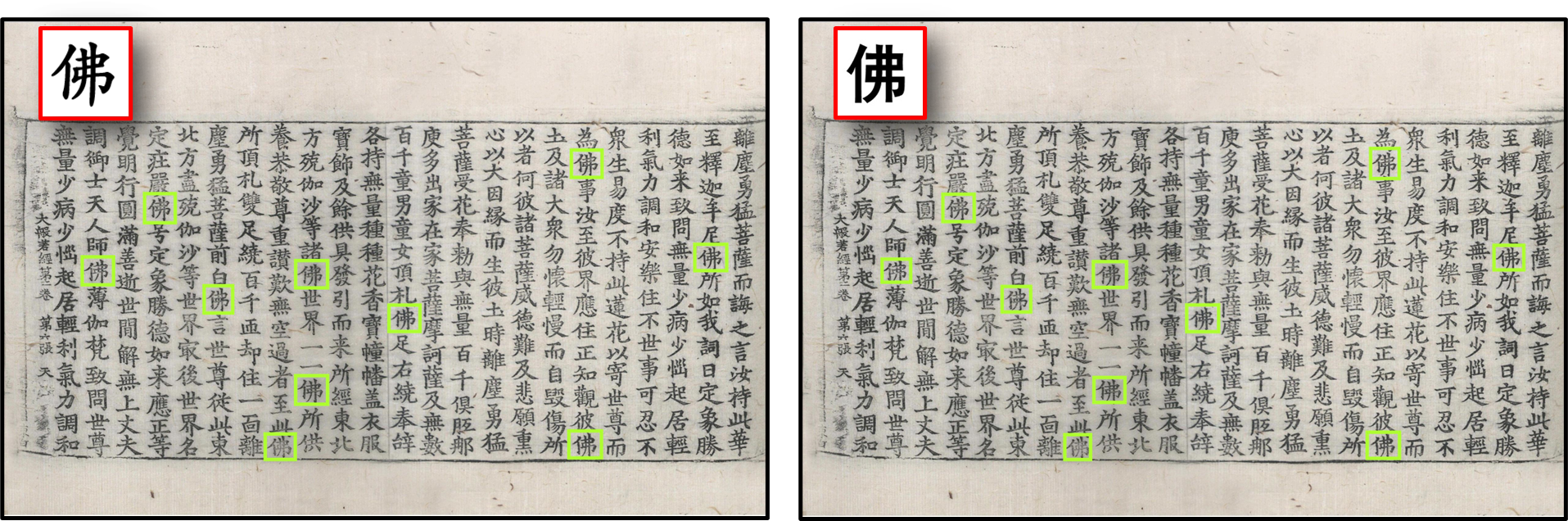}
	\end{center}
	\vspace{-0.5cm}
	\renewcommand{\figurename}{\textbf{Fig.}}
	\caption{Exemplary outcomes utilizing the TKH dataset are presented.}
	\label{fig:TKHFEATURE}
	\vspace{-0.3cm}
\end{figure}

In practical applications, character images with distinct glyph styles may be considered as support images for the spotting procedure. Quantitative analysis of this scenario is challenging; therefore, a qualitative assessment is provided in \cref{fig:TKHFEATURE}. The red box delineates the support image, while all predicted boxes are depicted in green. Within \cref{fig:TKHFEATURE},  it has been noted that using a support image with a different character style from the query image can produce favorable outcomes, thereby illustrating the model's generalization capacity and transferability. 

\begin{figure}[!h]
	\begin{center}
		\includegraphics[clip,width=8.5cm]{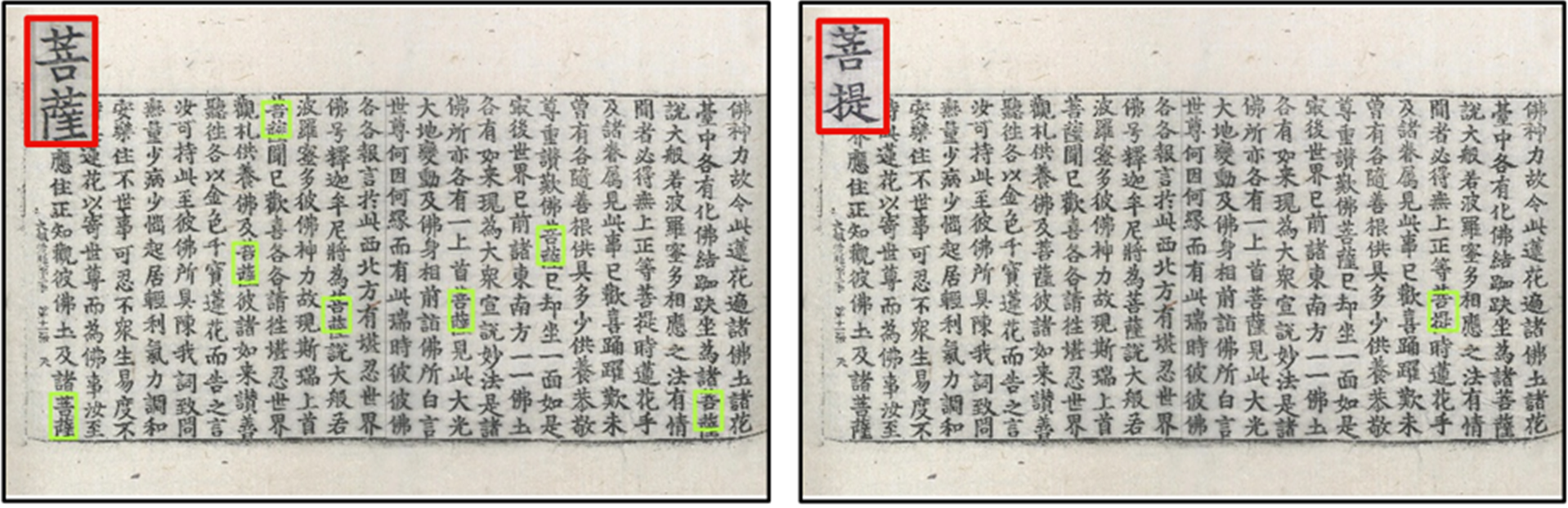}
	\end{center}
	\vspace{-0.5cm}
	\renewcommand{\figurename}{\textbf{Fig.}}
	\caption{Visualization of the VGTS quality in spotting novel combination characters.}
	\label{fig:WORDS}
	\vspace{-0.3cm}
\end{figure}

\textbf{Spotting Novel Combination Characters.} 
The proposed approach affords the flexibility to manage diverse and evolving application scenarios by enabling the spotting of novel classes. Specifically, we define a ``word" as a combination of characters belonging to a novel class when a single character sample serves as the support image during training. The extensive annotation required to determine which characters can constitute a word renders quantitative experimental comparisons challenging. Consequently, we present qualitative experimental results that illustrate the proposed approach's effectiveness. Impressively, our method delivers precise spotting performance for combination characters. \cref{fig:WORDS} showcase the performance of the proposed method on combination characters, emphasizing its capacity to accurately spot novel classes. It is essential to acknowledge that users must manually add unseen characters to the support image gallery for novel character classes, as our approach cannot automatically detect unknown categories in query images.

\begin{figure}[!t]
	\begin{center}
		\includegraphics[clip,width=8cm]{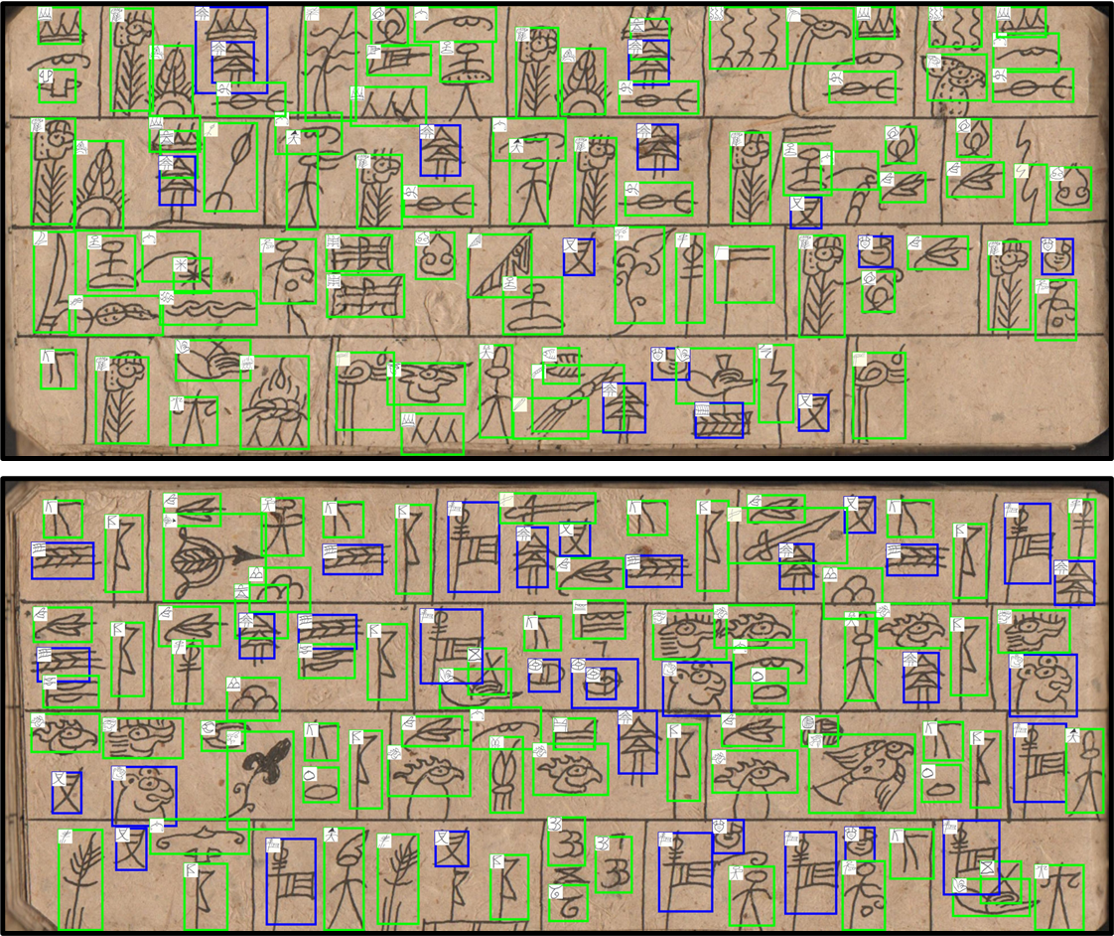}
	\end{center}
	\vspace{-0.5cm}
	\renewcommand{\figurename}{\textbf{Fig.}}
	\caption{Visualization results on the DBH datasets. \textcolor{green}{Green} boxes denote categories encountered during training (\textit{`Base'}), \textcolor{blue}{blue} boxes represent categories that are not present during the training process (\textit{`Novel'}). }
	\label{fig:DBHALL}
	\vspace{-0.3cm}
\end{figure}

\begin{figure}[!t]
	\begin{center}
		\includegraphics[clip,width=8.5cm]{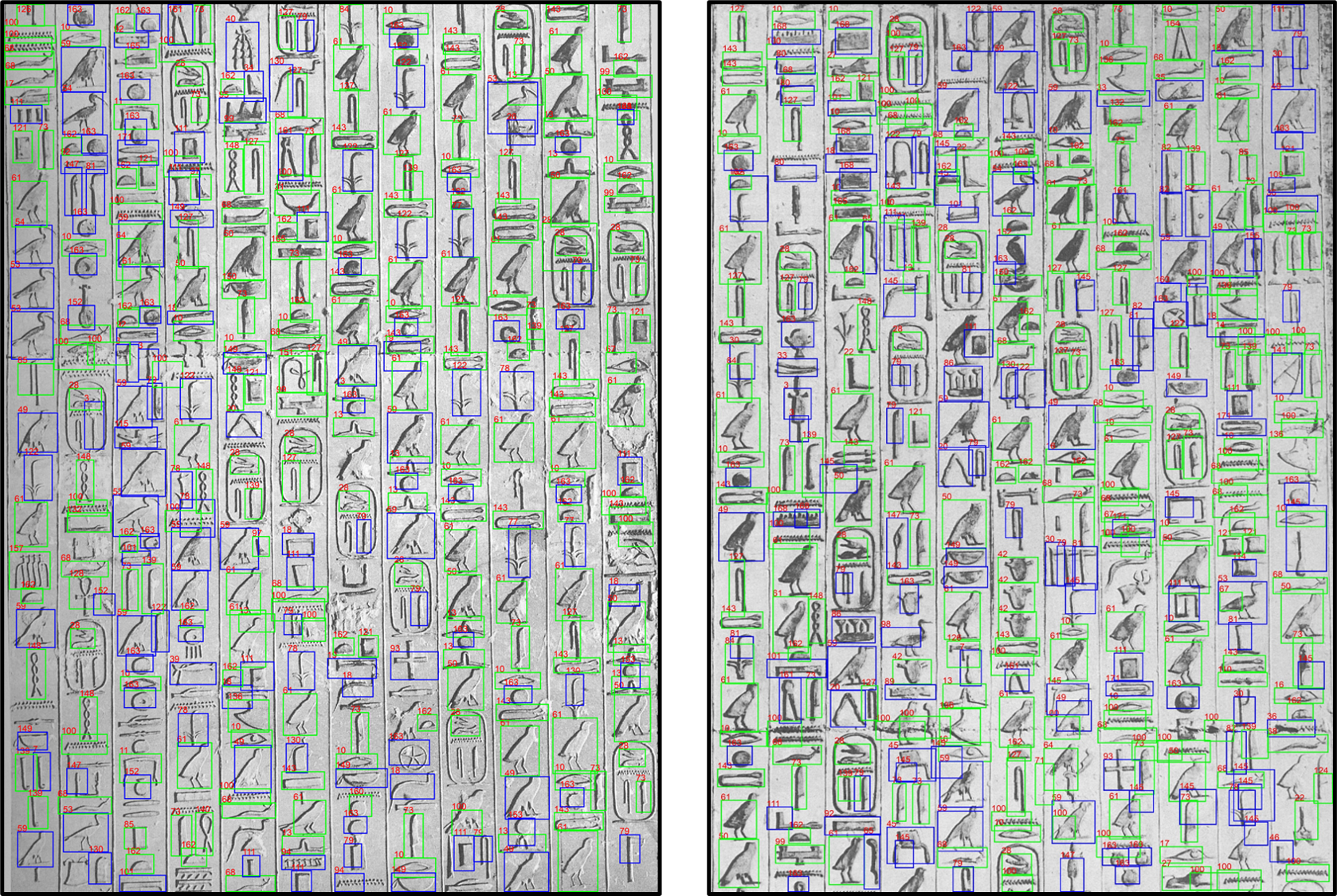}
	\end{center}
	\vspace{-0.5cm}
	\renewcommand{\figurename}{\textbf{Fig.}}
	\caption{Visualization results on the EGY datasets. \textcolor{green}{Green} boxes denote categories encountered during training (\textit{`Base'}), \textcolor{blue}{blue} boxes represent categories that are not present during the training process (\textit{`Novel'}). }
	\label{fig:EGYALL}
	\vspace{-0.3cm}
\end{figure}

\textbf{Spotting All Characters.} 
\cref{fig:DBHALL,fig:EGYALL} offer a glimpse of VGTS's visualization results on the DBH and EGY datasets and show its adeptness in identifying symbols from `Novel’ categories. Categories not part of the training are marked with blue boxes, while those that have been trained are highlighted with green boxes. VGTS possesses the capability to swiftly assist historians in cataloging character categories, this greatly benefits the decryption and study of historical manuscripts.

\subsection{Ablation Study}
\subsubsection{Initialization of the feature extractor}
When confronted with a limited amount of historical manuscript data, training the network from scratch proves to be a daunting task. As evidenced by the experimental results presented in \cref{tab:ABFE}, it becomes clear that the ResNet-50 backbone pre-trained on ImageNet (R50-ImgNet) delivers the best performance. Historical manuscripts in various languages may display unique patterns of texture features. We also tested the ResNet-50 pre-trained on the SynthText dataset \citep{sythtext} (R50-Synth) as the backbone, which similarly achieved high performance. Conversely, utilizing the ResNet-101 architecture did not yield superior results. Based on experimental observations, we opted for R50-ImgNet as the backbone for our model. Additionally, if we freeze the backbone to prevent feature learning, the model still performs quite well, achieving a 90.81$\%$ mAP on DBH Novel, as seen in \cref{tab:frozen}.
\begin{table}[!t]
	\renewcommand{\arraystretch}{1.1}
	\setlength{\tabcolsep}{4pt}
	\caption{\\ Ablation study of feature extractor and test on the novel categories of the DBH dataset. The best results are highlighted in \textbf{bold}.}
	\vspace{-0.5cm}
	\begin{center}
		{
			\begin{tabular*}{\linewidth}{@{}@{\extracolsep{\fill}}lcccccc@{}}
				\hline
				\rule{0pt}{3pt} \multirow{2}{*}{Component}  & \multicolumn{3}{c}{Novel} & \multicolumn{3}{c}{Base}       \\
				\cmidrule(lr){2-4} \cmidrule(lr){5-7}
				~                          & mAP & Recall & F1 & mAP & Recall & F1 \rule{0pt}{3pt}       \\	
				\hline
				\rule{0pt}{3pt} R50-ImgNet &\textbf{99.85}&\textbf{100.0}&\textbf{99.92}&\textbf{91.74}&\textbf{95.17}&\textbf{93.42} \rule{0pt}{3pt}\\
				\rule{0pt}{3pt} R50-Synth  &99.47&98.64&99.05&88.58&93.42&90.94 \rule{0pt}{3pt}\\
				\rule{0pt}{3pt} R101-ImgNet&97.03&99.29&98.15&86.30&92.88&89.47 \rule{0pt}{3pt}\\				
				\hline
		\end{tabular*}}
	\end{center}
	\vspace{-0.5cm}
	\label{tab:ABFE}
\end{table}
\begin{table}[!t]
	\renewcommand{\arraystretch}{1.1}
	\setlength{\tabcolsep}{4pt}
	\caption{\\ Ablation study of freezing backbone and test on the novel categories of the DBH dataset. The best results are highlighted in \textbf{bold}.}
	\vspace{-0.5cm}
	\begin{center}
		{
			\begin{tabular*}{\linewidth}{@{}@{\extracolsep{\fill}}lcccccc@{}}
				\hline
				\rule{0pt}{3pt} \multirow{2}{*}{Strategy}  & \multicolumn{3}{c}{Novel} & \multicolumn{3}{c}{Base}       \\
				\cmidrule(lr){2-4} \cmidrule(lr){5-7}
				~            &mAP           &Recall      &F1 &mAP           &Recall &F1\rule{0pt}{3pt}       \\	
				\hline
				\rule{0pt}{3pt} Unfreezing   &\textbf{99.85}&\textbf{100.0}&\textbf{99.92} &\textbf{91.74}  &\textbf{95.17} &\textbf{93.42} \rule{0pt}{3pt}\\
				\rule{0pt}{3pt} Freezing     &90.81         &98.21         &94.37 &75.75           &82.16&78.82 \rule{0pt}{3pt}\\			
				\hline
		\end{tabular*}}
	\end{center}
	\vspace{-0.5cm}
	\label{tab:frozen}
\end{table}
\begin{table}[!t]
	\renewcommand{\arraystretch}{1.1}
	\setlength{\tabcolsep}{3pt}
	\caption{\\ Performance comparison on different attention design methods of VGTS on the DBH dataset. The best results are highlighted in \textbf{bold}.}
	\vspace{-0.5cm}
	\begin{center}
		{
			\begin{tabular*}{\hsize}{@{}@{\extracolsep{\fill}}lcccccc@{}}
				\hline
				\rule{0pt}{3pt} \multirow{2}*{Configuration}  & \multicolumn{3}{c}{Novel} & \multicolumn{3}{c}{Base}       \\
				\cmidrule(lr){2-4} \cmidrule(lr){5-7}
				\rule{0pt}{3pt}~                      & mAP & Recall &F1 & mAP & Recall &F1\rule{0pt}{3pt}       \\	
				\hline
				\rule{0pt}{3pt} Support-First Att.         &\textbf{99.85}&\textbf{100.0}&\textbf{99.92}&\textbf{91.74}&\textbf{95.17}& \textbf{93.42}\rule{0pt}{3pt}\\
				\rule{0pt}{3pt} Query-First Att.   &98.73 &98.93 &98.83 &86.76 &89.69 & 88.20\rule{0pt}{3pt}\\
				\rule{0pt}{3pt} No Support Att.    &99.21 &\textbf{100.0} & 99.60 &87.89 &89.77 & 88.82\rule{0pt}{3pt}\\
				\rule{0pt}{3pt} No Query Att.      &96.10 &97.86 &96.97 &89.70 &93.45 & 91.54\rule{0pt}{3pt}\\	 
				\rule{0pt}{3pt} No Dual Att.       &96.65 &99.29 &97.95 &83.80 &87.64 & 85.68\rule{0pt}{3pt}\\
				\hline
			\end{tabular*}
		}
	\end{center}
	\vspace{-0.5cm}
	\label{tab:AB}
\end{table}

\begin{figure*}[!t]
	\begin{center}
		\includegraphics[clip,width=18cm]{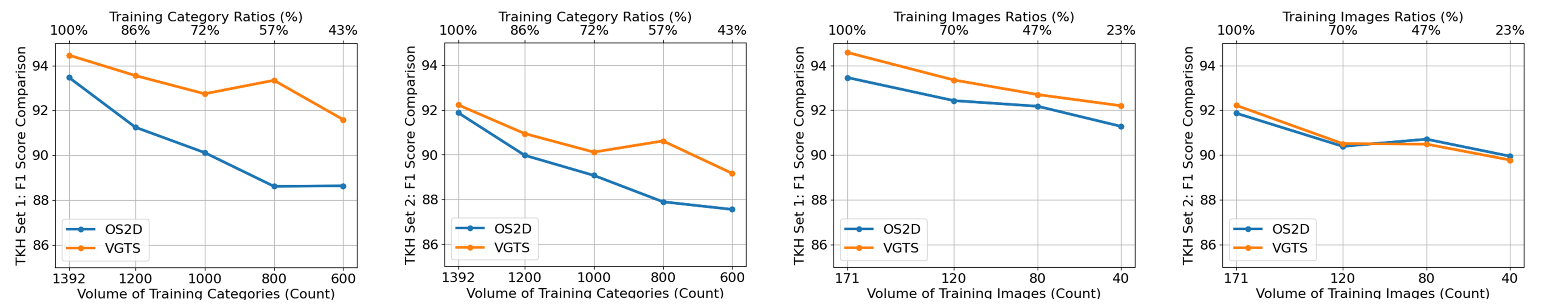}
	\end{center}
	\vspace{-0.2cm}
	\renewcommand{\figurename}{\textbf{Fig.}}
	\caption{The experiment of subsampling TKH categories during training, and testing on TKH Novel.}
	\label{fig:c_set1}
	\vspace{-0.3cm}
\end{figure*}
\begin{table*}[!t]
	\tabcolsep=0.1cm
	\renewcommand{\arraystretch}{1.1}
	\setlength{\tabcolsep}{3pt}
	\caption{\\ For testing the spotting loss function, we try several different settings of the loss function in the same architecture. We show the performance on the four datasets with \textbf{novel} class. The best results are highlighted in \textbf{bold}.}
	\vspace{-0.5cm}
	\begin{center}
		{
			\begin{tabular*}{\hsize}{@{}@{\extracolsep{\fill}}lccccccccccccccc@{}}
		\hline
				\rule{0pt}{3pt} \multirow{2}*{Loss Function}  & \multicolumn{3}{c}{DBH} & \multicolumn{3}{c}{VML}& \multicolumn{3}{c}{TKH Set 1} & \multicolumn{3}{c}{TKH Set 2} & \multicolumn{3}{c}{NC}       \\
				\cmidrule(lr){2-4} \cmidrule(lr){5-7} \cmidrule(lr){8-10} \cmidrule(lr){11-13} \cmidrule(lr){14-16}
				~              & mAP & Recall &F1 & mAP & Recall &F1 & mAP & Recall &F1 & mAP & Recall &F1 & mAP & Recall &F1\rule{0pt}{3pt}     \\	
				\hline
				\rule{0pt}{3pt} Triplet         &79.25&86.79&82.85&60.00&86.86&70.97&71.28&97.75&82.44&62.19&95.32&75.27&88.83&95.41&92.00 \rule{0pt}{3pt}\\
				\rule{0pt}{3pt} Contrastive     &92.11&97.14&94.56&98.23&98.73&98.48&75.79&99.05&85.87&69.43&97.48&81.10&90.60&94.50&92.51 \rule{0pt}{3pt}\\
				\rule{0pt}{3pt} RL              &99.47&99.64&99.55&98.73&99.58&99.15&90.05&99.60&94.58&86.99&98.13&92.22&90.77&95.72&93.17 \rule{0pt}{3pt}\\
				\rule{0pt}{3pt} Torus (Ours)                   &\textbf{99.85}&\textbf{100.0}&\textbf{99.92}&\textbf{100.0}&\textbf{100.0}&\textbf{100.0}&\textbf{90.20}&\textbf{99.65}&\textbf{94.69}&\textbf{88.06}&\textbf{99.52}&\textbf{93.44}&\textbf{92.14}&\textbf{96.79}&\textbf{94.41} \rule{0pt}{3pt}\\	 
		\hline
			\end{tabular*}
		}
	\end{center}
	\vspace{-0.5cm}
	\label{tab:loss}
\end{table*}
\subsubsection{Influence of Attention Design}
To empirically validate the effectiveness of our proposed framework, we conduct a comprehensive ablation study on the dual spatial attention block. Specifically, we perform one of the following operations at a time: (1) interchanging the order of the two spatial attention operations, and (2) removing one of the spatial attention operations. We evaluate all the models on the DBH dataset, and the results are presented in \cref{tab:AB}. The human cognitive process typically involves examining the support image first and then seeking relevant similar regions in the query image. By altering the order of spatial attention in VGTS, with query spatial attention preceding support spatial attention, we observe a decrease in mAP scores by approximately 0.7$\%$ and 3$\%$ for the novel and base categories, respectively. The experimental results attest to the effectiveness of the proposed module sequence. Eliminating the spatial attention for support images resulted in a decrease of about 1.7$\%$ in mAP scores for the base category, and removing the spatial attention for query images led to approximately a 3.4$\%$ decrease in mAP scores for the novel category, providing evidence that the spatial attention for both support and query images plays a crucial role in achieving stable and significant improvements.
\subsubsection{Subsampling Experiment}
In contrast to widely-used object detection datasets such as COCO \citep{COCO} and VOC \citep{VOC}, historical manuscript images present unique challenges, including smaller sample sizes, a larger number of categories, and considerable similarities between categories. To assess the efficacy of our approach in low-resource scenarios, we conduct experiments by varying the number of base categories in the training set while maintaining the number of input images constant. 
\cref{fig:c_set1} presents a performance comparison between our proposed method and OS2D on the TKH dataset, showcasing results for varying numbers of training set categories, ranging from 600 to 1392. As illustrated in \cref{fig:c_set1}, the number of categories in the training set significantly influences the performance outcomes. Specifically, when the number of participating categories is reduced by 792, there is a noticeable impact on the F1 scores. For OS2D, this reduction leads to a decrease of 4.83\% on the TKH Set 1 and 4.32\% on the TKH Set 2. In contrast, for our VGTS method, the F1 scores decline by 2.87\% on the TKH Set 1 and 3.06\% on the TKH Set 2.

To investigate the influence of the number of input images on the model's performance, we further evaluate the model on training sets comprising 40, 80, 120, and 171 images, corresponding to a total of 1104, 1232, 1320, and 1942 categories, respectively. The findings depicted in \cref{fig:c_set1} indicate a positive correlation between the number of input images and model performance, as evidenced by higher F1 scores with larger image datasets. Notably, our method demonstrates robustness even in low-resource conditions, achieving F1 scores of over 92.2\% and 89.78\% on the TKH Set 1 and TKH Set 2 datasets, respectively, when trained with just 40 images.
\subsubsection{Selection of Loss Function}
The loss function comprises both localization and spotting components. For localization, we employ the smooth $L_{1}$ loss, while for the spotting loss, we assess the effectiveness of our proposed new loss by testing several different configurations of the loss function within the same architecture. Our results in \cref{tab:loss} demonstrate that the triplet loss function does not yield substantial performance improvements, whereas the contrastive loss function surpasses the triplet loss function. The ranked list loss, which can adaptively balance positive and negative samples, achieves commendable performance. However, the proposed torus loss function outperforms all other loss functions. This can be attributed to its ability to focus on challenging samples in the margin gap and balance positive and negative samples, which facilitates learning distance metrics and enhances text spotting in low-resource scenarios. The experimental results on all four datasets using the torus loss function corroborate its superiority compared to other loss functions, thus validating the effectiveness of our proposed approach.
\begin{table}[!t]
	\renewcommand{\arraystretch}{1.1}
	\setlength{\tabcolsep}{3pt}
	\caption{\\ Cross-domain performance measures of our method when training is done on different datasets and testing on DBH Novel.}
	\vspace{-0.5cm}
	\begin{center}
		{
			\begin{tabular*}{\linewidth}{@{}@{\extracolsep{\fill}}lccccc@{}}
				\hline
				\rule{0pt}{3pt} \multirow{2}*{Metric} & \multicolumn{5}{c}{Training Dataset}\\
				\cmidrule(lr){2-6}
				~ & DBH & EGY &VML & TKH & NC \rule{0pt}{3pt} \\	
				\hline
				\rule{0pt}{3pt} mAP              &99.85&97.79&98.44&92.90&58.23 \rule{0pt}{3pt}   \\
				\rule{0pt}{3pt} Recall           &100.0&99.64&99.29&98.57&61.79 \rule{0pt}{3pt}   \\
				\rule{0pt}{3pt} F1               &99.92&98.71&98.86&95.65&59.96 \rule{0pt}{3pt}   \\		 
				\hline
		\end{tabular*}}
	\end{center}
	\vspace{-0.5cm}
	\label{tab:CROSS}
\end{table}
\subsubsection{Cross-domain Performance}
We have conducted a comprehensive evaluation of the cross-domain performance of our proposed model by training it on various datasets and selecting the DBH Novel dataset for testing purposes. The experimental results are presented in \cref{tab:CROSS}, demonstrating the effectiveness of our proposed model. For example, when the model is trained on the VML dataset, the DBH Novel dataset can be considered as a novel class. However, the model's optimal performance is obtained when it is trained and tested on the same dataset, as the data distributions are similar. Nevertheless, cross-domain training may result in a decline in performance, particularly when the data types differ significantly, and a limited number of samples are used in the training process.

\section{Conclusions}
Existing object detection methods struggle to identify classes that are not part of the training, particularly in low-resource scenarios. Addressing this challenge, this paper presents a novel approach for page-level text spotting in historical manuscripts using deep learning techniques, referred to as VGTS, that effectively addresses the low-resource open-set problem. The proposed method leverages dual spatial attention and correlation matching to align support and query images, closely emulating the human cognitive process. The approach enables the spotting of new classes without additional training and is particularly suitable for low-resource scenarios. To address example imbalance, we introduce a novel loss function called torus loss that makes the embedding space of distance metric more discriminative. Our experimental results demonstrate that our approach significantly outperforms existing methods, thereby offering promising applications in this field. 
\section*{Acknowledgments}
This work was jointly supported by the National Natural Science Foundation of China under Grant No. 62176091, the National Key Research and Development Program of China under Grant No. 2020AAA0107903.

\section*{Annex. The Challenges of Dongba Script Research.}
The Dongba script, a unique hieroglyphic language developed by the Naxi minority's ancestors in China, represents a vital part of cultural heritage and historical linguistics. Despite its importance, researchers in the field face several formidable challenges in deciphering and cataloging this ancient script. The challenges encountered in the study of Dongba script can be summarized as follows:

\begin{figure}[!h]
	\begin{center}
		\includegraphics[clip,width=8.5cm]{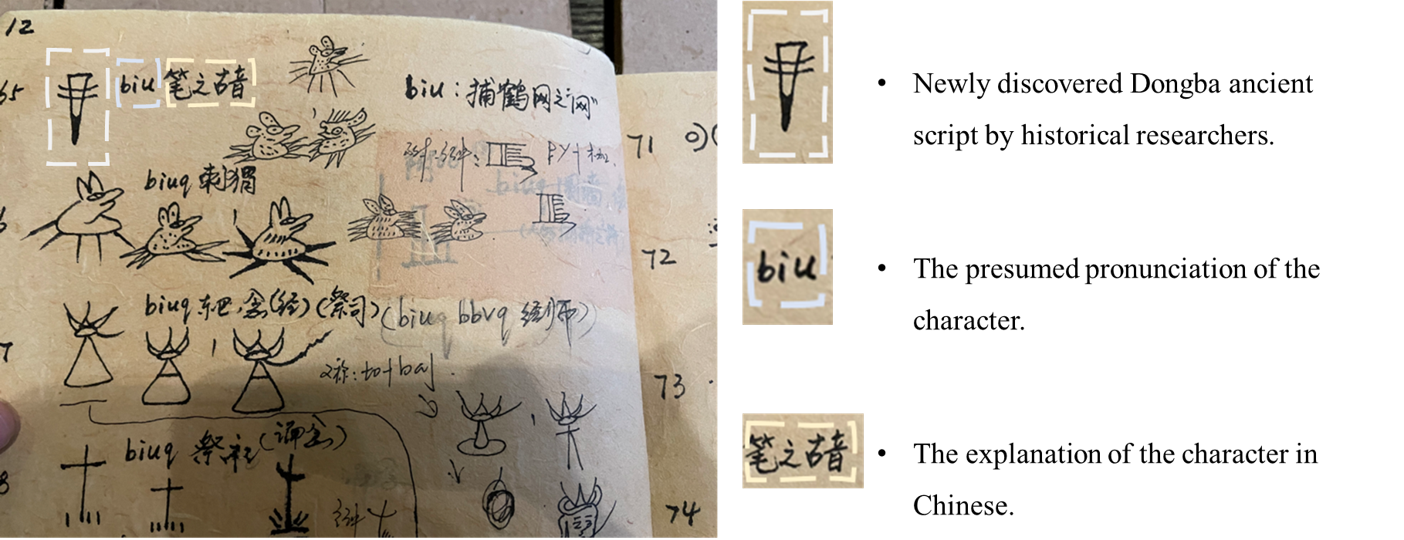}
	\end{center}
	\caption{Manuscript notes of a historian studying Dongba texts: newly discovered characters categorized and annotated.}
	\label{fig:DBR}
	\vspace{-0.5cm}
\end{figure}
\begin{enumerate}
	\item \textbf{Resource Limitations and Manual Deciphering:} Given the scarcity of resources, researchers often engage in manual identification and documentation of new symbols in Dongba manuscripts. This labor-intensive process demands high levels of accuracy and attention to detail, as illustrated in \cref{fig:DBR}, which shows the painstaking process a scholar undergoes when documenting insights from Dongba manuscripts.
	
	\item \textbf{Human Factor in Interpretation:} The manual nature of this work introduces the risk of inefficiencies, potential misinterpretations, and inaccuracies. These human factors can significantly impact the reliability of the research findings.
	
	\item \textbf{Discovery of Novel Characters:} One of the most significant challenges in Dongba script research is the ongoing discovery of new characters and markings. This evolving nature of the script continually adds complexity to the research process.
	
	\item \textbf{Technological Limitations:} Traditional object detection methods, typically designed under a closed-set assumption, struggle to recognize these newly discovered characters. Adapting these methods requires constant model retraining, which can be resource-intensive and may detract from the primary research focus.
\end{enumerate}


\bibliographystyle{elsarticle-harv}
\bibliography{VGTS}



\end{document}